\documentclass{article}

\usepackage{arxiv}

\usepackage[utf8]{inputenc} % allow utf-8 input
\usepackage[T1]{fontenc}    % use 8-bit T1 fonts
\usepackage{hyperref}       % hyperlinks
\usepackage{url}            % simple URL typesetting
\usepackage{booktabs}       % professional-quality tables
\usepackage{amsfonts}       % blackboard math symbols
\usepackage{algorithm}% http://ctan.org/pkg/algorithms
\usepackage{algpseudocode}%[noend] http://ctan.org/pkg/algorithmicx
\usepackage{hyperref}
\usepackage[textsize=tiny]{todonotes}
\usepackage{graphicx}
\usepackage{colortbl}
\usepackage{eurosym}
\usepackage{boldline}
\usepackage{float}
\usepackage{xcolor}
\usepackage{calc}
\usepackage{nccmath}
\usepackage{xspace}
\usepackage{setspace}
\usepackage{romannum}
\usepackage{soul}
\usepackage[export]{adjustbox}
\usepackage[caption=false]{subfig}
\usepackage{nicefrac}       % compact symbols for 1/2, etc.
\usepackage{microtype}      % microtypography
\usepackage{lipsum}
\usepackage{rotating,tabularx}

\newcolumntype{?}{!{\vrule width 2pt}}
\usepackage{multirow}
\usepackage{amsmath}
\usepackage{pdflscape}
\usepackage{afterpage}
\usepackage{capt-of}
\title{Covariance Matrix Adaptation Greedy Search Applied to Water Distribution System Optimization}

\author{
  Mehdi Neshat \\
  Optimization and Logistics Group\\
  School of Computer Science\\
  The University of Adelaide\\
   Australia \\
  \texttt{mehdi.neshat@adelaide.edu.au} \\
   \And
 Bradley Alexander \\
  Optimization and Logistics Group\\
  School of Computer Science\\
  The University of Adelaide\\
   Australia \\
  \texttt{bradley.alexander@adelaide.edu.au} \\
  \And
 Angus Simpson \\
  School of Civil, Environmental and Mining Engineering\\
  The University of Adelaide\\
  Australia \\
  \texttt{angus.simpson@adelaide.edu.au} \\
  }

\begin{document}
\maketitle

\begin{abstract}
Water distribution system design is a challenging optimisation problem with a high number of search dimensions and constraints. 
In this way, Evolutionary Algorithms (EAs) have been widely applied to optimise WDS to minimise cost subject whilst meeting pressure constraints. This paper proposes a new hybrid evolutionary framework that consists of three distinct phases. The first phase applied CMA-ES, a robust adaptive meta-heuristic for continuous optimisation. This is followed by an upward-greedy search phase to remove pressure violations. Finally, a downward greedy search phase is used to reduce oversized pipes. 
To assess the effectiveness of the hybrid method, it was applied to five well-known WDSs case studies. The results reveal that the new framework outperforms CMA-ES by itself and other previously applied heuristics on most benchmarks in terms of both optimisation speed and network cost.
\end{abstract}


\section{Introduction}

Water distribution systems (WDSs) are expensive to construct~\cite{alperovits1977design} and difficult to modify once in place. Careful design of WDS systems can lead to significant cost savings. As a consequence, optimisation of WDS design is a long-standing topic for research. 

The optimisaton problem is framed as a problem of minimising the cost of WDS construction subject to given constraints in head-pressure at each WDS network node. The optimisation problem is challenging because relationships between individual pipe sizes and pressure at nodes are non-linear. In addition, the search problem, for non-trivial networks, is non-convex with many local minima in the cost function. Moreover, the problem constraints are quite complex, with detailed requirements for head pressures and, commonly, restriction of pipe sizes to discrete values.

During the last two decades, a wide variety of Evolutionary Algorithms (EAs) has been applied to the problem of optimizing the design of WDSs. EAs offer flexibility in their design parameters and work robustly in multi-modal, nonlinear and non-convex fitness spaces compared with traditional optimization methods such as linear~\cite{fujiwara1987modified} and nonlinear programming~\cite{fujiwara1990two}. 
In early work with EAs good results  have been achieved by standard genetic algorithms (GA)~\cite{savic1997genetic,wu2001using}. Later work modified a standard GAs with an additional heuristic selection phase  (the Prescreened Heuristic Sampling Method (PHSM)) for initialising the population of the GA~\cite{bi2015improved};  Other EAs applied include Ant Colony Optimization (ACO)\cite{maier2003ant,geem2009particle} and modified versions of ACO including the Max-
Min Ant Systems \cite{zecchin2006application,zecchin2007ant} and the adaptive-convergence-trajectory ACO~\cite{zheng2017adaptive}. Other heuristic approaches have included, Simulated Annealing (SA)~\cite{cunha1999water}, Scatter Search~\cite{lin2007scatter}, the Shuffled Frog Leaping algorithm~\cite{eusuff2003optimization}, Particle Swarm Optimisation (PSO)~\cite{montalvo2008particle,aghdam2014design} and heuristics embedding Differential Evolution (DE)~\cite{suribabu2010differential,zheng2012performance,zheng2012self}. Still other approaches have combined heuristics including: DE and linear-programming~\cite{zheng2013coupled};
DE and non-linear-programming~\cite{zheng2011combined}; and PSO and DE~\cite{sedki2012hybrid}.

Both hybridization and adaptation have  considered as a proper alternative approach to developing the optimization results. Zheng et al. ~\cite{zheng2017adaptive} propose an adaptive convergence trajectory controlled idea for improving the diversification and intensification of ACO. As a result, in two large-scale networks, better solutions are found compared with previous methods. In 2013 \cite{zheng2012self}, a self-adaptive DE with new convergence criteria represented and except the BN case study, its performance is acceptable. A hybridization of DE and Linear Programming can be an excellent example of how one evolutionary algorithm abilities are strengthened by a traditional optimization method that leads to modify the convergence rate and accuracy, especially the Hanoi network results are appreciable~\cite{zheng2013coupled}. The PSO-DE is another hybrid method that benefits from the advantages of both PSO and DE like the ability to attain the optimum solution and computational efficiency \cite{sedki2012hybrid}; however, its performance did not evaluate by big networks.

Despite achieving the substantial outcomes by various rang of EAs, their performances have been considerably affected by the initialising of algorithm parameters control in terms of efficiency and robustness. Therefore, adjusting these parameters which are done by trial and error most of the time depends on optimisation problem characteristics. However, in the majority of cases, it can be extremely costly in terms of computational budgets. This issue is the main reason why recently there is an upward trend of applying adaptive and self-adaptive EAs.
CMA-ES~\cite{hansen2004evaluating} is a self-adaptive, global search algorithm designed for searching spaces consisting of many continuous variables. CMA-ES is one of the fastest and versatile meta-heuristic search algorithms. Variants of CMA-ES often lead black-box optimisation competitions and also it has shown to perform well in noisy, non-convex and non-separable search spaces \cite{Skvorc:2019:GBO:3319619.3321996}. 

A CMA-ES search process in $n$ dimensions works by adapting a $n\times n$ covariance-matrix $C$ which defines the shape and orientation of a Gaussian distribution in the search space and a vector $x$ that describes the location of the centre of the distribution. The distribution is used to bias the location of the next set of sampled individual solutions in the search space. The relative performance of these individual samples is used to update both $C$ and $x$. This process of sampling and adaptation continues until search converges, or a fixed number of iterations have expired.

Hence,  this research demonstrates the application and comparison of three EAs like CMA-ES, Randomized Local Search (RLS) and 1+1EA for optimizing five case studies of WDSs in the single-objective domain, and more significantly, focuses on improving the CMA-ES performance through the incorporation of a non-derivative and gradient-free search techniques including Greedy Search (GS), which results in a hybrid evolutionary framework. According to the achieved results of five networks, The hybrid method can outperform other Evolutionary methods, and its findings of the large-scale networks show a suitable average performance which is more considerable compared with the previous studies.  
%---------------------------------------------------
\begin{table}
\caption{Software availability }\label{details_software}
\centering
\begin{tabular}{|l p{8cm}|}
\hlineB{4}
Name of the Software: & WDSOP \\ 
Version: & 1.00\\
Available from: & https://github.com/a1708192/WDSOP\\
Language:  &  Matlab\\
Supported System: & Windows, MacOS, Linux, Unix \\
Year first available: & 2019 \\
  \hlineB{4}
  \end{tabular}
\end{table}
%---------------------------------------------

This research represents an extendable, robust and portable optimization framework for minimizing the pipe cost of the WDSs called WDSOP. WDSOP is an adaptive hybrid evolutionary framework which is implemented by Matlab programming language. A modular programming design (structured programming) is applied that enables researchers to combine and modify solution components efficiently.  WDSOP has been evaluated for optimizing five realistic benchmarks in terms of the robustness and convergence rate. The only input variable of the framework is the water distribution network name, and other control parameters are autonomously adjusted based on the optimization process. Table \ref{details_software} shows the details of the proposed Matlab framework (WDSOP).   

The organization of the paper rest is as follows. Some preliminaries of methodologies are explained in section \ref{sec:methods}. Section \ref{sec:WDS-details} presents the details of applied case studies in this research. The experimental results and their analysis are outlined in section \ref{sec:results}. Finally, the conclusion and the directions of the future work are summarized.

\section{Methodology} \label{sec:methods}

\subsection{Covariance Matrix Adaptation Evolution Strategy (CMA-ES)}

 CMA-ES relies on three principal operations, selection, mutation, and recombination. Recombination and mutation are employed for exploration of the search space and creating genetic variations, whereas the operator of selection is for exploiting and converging to an optimal solution. The mutation operator plays a significant role in CMA-ES, which utilizes a multivariate Gaussian distribution. A thorough explanation of different selection operators exposes to \cite{beyer2002evolution}.
 
Indeed, CMA-ES can be a right candidate for exploring and exploiting as a deep local search which equips with a self-adaptive mechanism for setting a suitable vector of mutation step sizes($\sigma$) instead of having just one global mutation step size. This is because one step size cannot be very efficient in resolving high-dimensional problems. Applying a multivariate Gaussian distribution with the proper size of $\sigma$ and mean causes the acceptable convergence velocity and diversity \cite{hansen2004evaluating}. The covariance matrix is computed based on the differences in the mean values of two progressive generations. In which case, it expects that the current population includes sufficient information to estimate the correlations favourably. After calculating the covariance matrix, the rotation matrix will derive from the covariance matrix with regard to expanding the distribution of the multivariate Gaussian in the right direction of the global optimum. It can accomplish by conducting an eigen-decomposition of the covariance matrix to receive an orthogonal basis for the matrix \cite{hansen2014principled,hansen2006cma}.

 \subsection{Randomized Local Search(RLS) and 1+1EA}
Randomized Local Search (RLS) is the simplest single-based solution EAs. According to the practical results, sometimes applying the simple EAs can be more efficient than complicated approaches and also RLS can be a proper choice when the fitness function is a combinatorial optimization problem~\cite{neumann2007randomized}.
 RLS begins with a candidate solution ($x$) and provides in each iteration a new solution ($y$) by flipping one chosen variable of $x$ randomly. In the standard version of RLS, the mutation is done by a uniform distribution which leads to a non-curved and noisy local search, but we prefer to use a normally distributed mutation. The advantage of RLS is that in each iteration, just one pipe size of the network is changed. This attribute leads to approaching a near-optimal solution step by step; however, it can be so costly for a large-scale network. In the following, the pseudo-code of RLS can be seen by the Algorithm \ref{alg:RLS}.
%---------------------------------------------------
\begin{algorithm}
 \caption{The RLS}\label{alg:RLS}
  \begin{algorithmic}[1]
  \Procedure{ The RLS}{}
 \\ \textbf{Initialization}
  \State LB=Min(Diameters);UB= Max(Diameters) 
   \State $X_{iter} \in \{LB,UB \}^N$ uniformly at random \Comment{Generate first feasible design}
  \While{Stopping Criteria}
 \\ \textbf{Mutation}
  \State Create $Y_{iter}=X_{iter}$ independently for each $i \in \{1,2,...,N\}$
 \State $ Y_{iter}=N(\mu, \sigma= 0.5*(UB-LB)$) \Comment{Mutate one random variable of $Y_{iter}$by normally distributed random }
  %\State Handle the boundaries constraint\\ $Y_{iter}\in \{LB,UB \}^n$
  \\ \textbf{Selection}
  \If{($f(Y_{iter}) \le f(X_{iter} $))}
  \State $X_{iter+1}= Y_{iter}$
  \Else
  \State $X_{iter+1}= X_{iter}$
  \EndIf
  \EndWhile\
   \EndProcedure
  \end{algorithmic}
 \end{algorithm}
%----------------------- 
where $UB$ and $LB$ are the upper and lower bound of the variable, and also $n$ is the number of variables. 
Undoubtedly, after RLS, the most simple evolutionary algorithm is (1+1)EA because there is just one solution in each iteration and a standard bit mutation applies for providing a new solution with mutation probability $\frac{1}{N}$ that N is the number of variables. Its benefits like simplicity and performance make (1+1)EA one of the most attractive EAs which can often be generalized and extended to more complex EAs. As (1+1)EA performance can be better than complex EAs like CMA-ES in some cases \cite{neshat2018detailed}, in this investigation, fine-tuned mutation step size of (1+1)EA version is implemented for optimizing the WDSs and analyzing its pros and cons. Besides, since there is the probability that in one mutation, none of the variables is mutated, a substitute mutation strategy is considered, which is flipping the size of one pipe randomly at least. The pseudo-code of (1+1)EA can be indicated in the Algorithm \ref{alg:1+1EA}.

%1+1EA-----------------------------
\begin{algorithm}
\caption{The (1+1)EA}\label{alg:1+1EA}
\begin{algorithmic}[1]
\Procedure{ The (1+1)EA}{}
 \\ \textbf{Initialization}
\State LB=Min(Diameters);UB= Max(Diameters) 
\State $X_{iter} \in \{LB,UB \}^N$ uniformly at random \Comment{Generate first feasible design}
\While{Stopping Criteria}
\\ \textbf{Mutation}
\State Create $Y_{iter}=X_{iter}$ independently for each $i \in \{1,2,...,N\}$
\State Mutate each variable of $Y_{iter}(i)$ with probability $\frac{1}{N}$ (Normal random distribution: $\sigma=C$)
\If {Mutation Number= 0}
\State Mutate one variable of $Y_{iter}$ randomly
\EndIf
\\ \textbf{Selection}
\If{($f(Y_{iter}) \le f(X_{iter} $))}
 \State $X_{iter+1}= Y_{iter}$
\Else
 \State $X_{iter+1}= X_{iter}$
\EndIf
\EndWhile\
\EndProcedure
\end{algorithmic}
\end{algorithm}
%---------------------------------------------------
\subsection{Covariance Matrix Adaptation Greedy Search (CM$A_{ES}$-G$S_U$-G$S_D$)}
In this study, a hybrid Evolutionary scheme proposes that called CM$A_{ES}$-G$S_U$-G$S_D$ (Algorithm \ref{alg:CMAGS}). The hybridization scheme is involved in three distinguished levels. Applying a CMA-ES which has a pivotal role is the first step. The CMA-ES is a self-adaptive stochastic method, and when the fitness functions are nonlinear or non-convex, its performance can be competitive. The CMA-ES benefits the cooperation of covariance matrix adaptation that is constrained into a multivariate normal distribution. The purpose of the adaptation of covariance matrix signifies to approximate the inverse Hessian matrix, like a quasi-Newton method to the covariance matrix. The aim can be satisfied with the distribution of the search process to the contour lines of the fitness function. These characteristics of CMA-ES make a robust optimization method, especially in continuous search spaces. Although pipe diameters of the networks are discrete, participating in two different direct search techniques for modifying the results of CMA-ES is the primary motivation of the second part of the proposed hybrid framework.  CMA-ES is able to find the very cheap configurations of the network design compared with other EAs, but these cheap proposed layouts are not feasible in terms of the nodal pressure head constraints. Thus, for compensating the issue, an Upward Greedy Search ($GS_U$) technique (Algorithm~\ref{alg:UGS}) contributes to the CMA-ES.  

%----- Upward Greedy Search------------------------------
 
 \begin{algorithm}
 \caption{The Upward Greedy Search (Fixing up the  nodal pressure head violations)}\label{alg:UGS}
 \begin{algorithmic}[1]
 \Procedure{ The Upward Greedy Search}{}\\
  \textbf{Initialization}
  \State $\mathit{Initialize}~\tau$ \Comment{initialize Pressure head constraint}
  \State $Layout_{iter}$=Pipe-Network \Comment{Read Network data }
  \State $Nodal_{Pressure}$=$\mathit{Eval(Layout_{iter})}$ \Comment{Evaluate Network by EPANET }
  \State $N=Size(Layout_{iter})~and~M=Size(Nodal_{Pressure})$
  \State $Sum_{PV}=\sum_{i=1}^{M} (|\tau-Nodal_{Pressure}^{i}|)~ \forall~ 
   Nodal_{Pressure}^{i} <\tau$ \Comment{Calculate sum of the nodal pressure violation}
  \While{$Sum_{PV} > 0$}
  \State$NetPipe=Layout_{iter}$
  \While{$i\le N$}
  \State Increase the $i^{th}$Pipe diameter of the $Netpipe_{iter}$ based on the possible diameters
  \State $Nodal_{Pressure}$=$\mathit{Eval(Netpipe_{iter})}$ 
 \State $\mathit{Update}~Sum_{PV}$
  \State $Improvement_{rate}^{i}=\frac{ \sum_{i=1}^{M}\triangle PV_{i}}{\sum_{i=1}^N\triangle PipeCost_{i}}$
 \EndWhile
 \State $Layout_{iter}$=Max($Improvement_{rate}$) \Comment{ Choose the best design based on Improvement rate}
    \EndWhile\
%\State \textbf{end while}
 \EndProcedure
 \end{algorithmic}
 \end{algorithm}
%-------------------------------------------------
The $GS_U$ enhances the infeasibility amount of the CMA-ES achieved solutions and pushes up the infeasible layouts toward the feasible area by increasing the discrete size of pipe diameters based on a greedy selection of those solutions with the largest reduction in the sum of pressure violations for the least cost. The maximization problem can be stated mathematically as: 
%----------------------------------------
\begin{equation}
\begin{split}
Argmax \to  f(\Theta)=(\frac{ \sum_{i=1}^{M}\triangle PV_{i}}{\sum_{j=1}^N\triangle PipeCost_{j}}) 
\\\forall i=\{1,...,M\} /&j=\{1,...,N\} \\
Subject-to: \triangle PV_{i} \le 0 ,\triangle PipeCost_{j}\ge 0
 \end{split}
\end{equation}
%-------------------------------------------
 Where a layout $\Theta$ can be defined as a set of sequential pipe diameters, so $\Theta=\{D_1,D_2,...,D_N\}$ that N is the number of pipes and M is the number of network nodes.  $f(\Theta)$ mentions the feasibility of the network function, which should be maximized by the $GS_U$. This greedy heuristic search method is able to guarantee to produce a feasible design based on the constraints, and also it will yield the locally optimal design in a reasonable runtime. The procedure of the $GS_U$ is shown in the Algorithm \ref{alg:UGS}.

Despite all positive points of $GS_U$, sometimes its proposed solutions require improvements because of the greedy selection behaviours without looking at the future or past situations. Therefore, the third phase is proposed to reduce the extra cost of the some of the obtained solutions. This part is made up of another Greedy Search idea.

%------penalty figure------------------------
 \begin{figure} 
 \centering
 \includegraphics[ width=0.5\textwidth,height=3.0in]
 {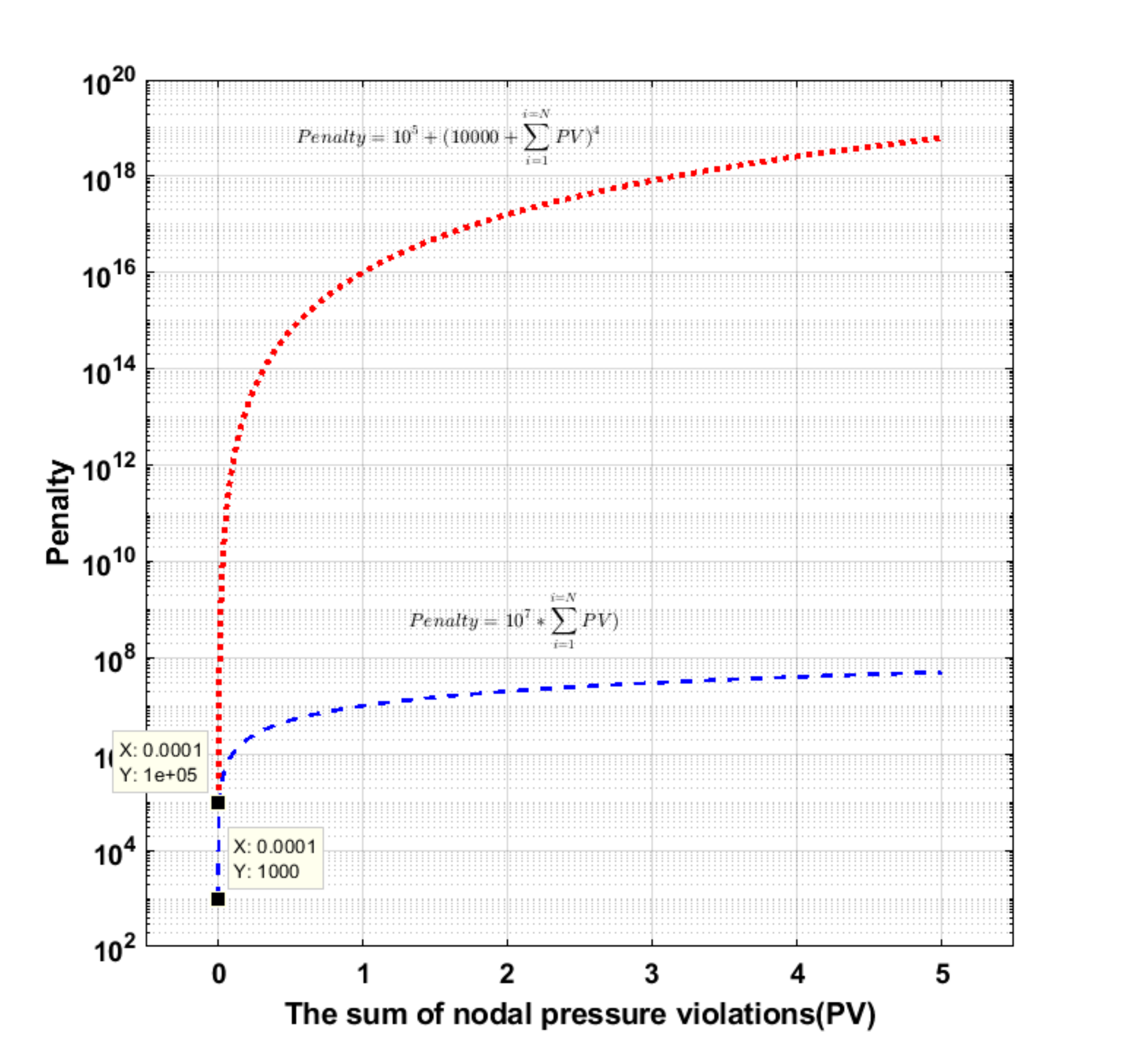}
 \caption{ Two applied penalty functions for handling the pressure violations. The red line is a severe penalty function.  } \label{fig:penalty_function}
 \end{figure}
   %----------------------------------------
%---------------------------------
\begin{figure*}[t]%[H]
\centering

\subfloat[]{
\includegraphics[clip,width=0.45\columnwidth]{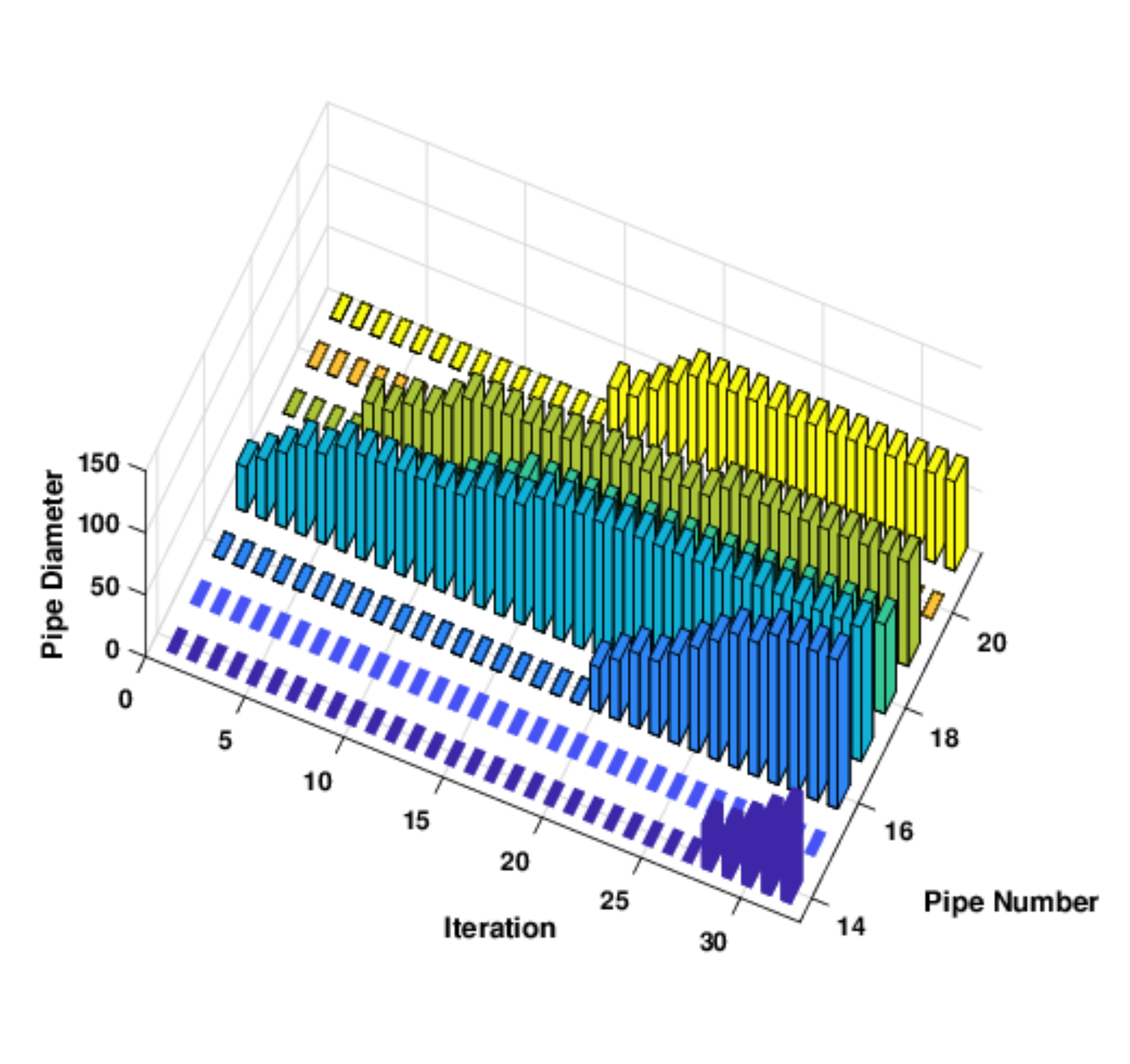}}
\subfloat[]{
\includegraphics[clip,width=0.45\columnwidth]{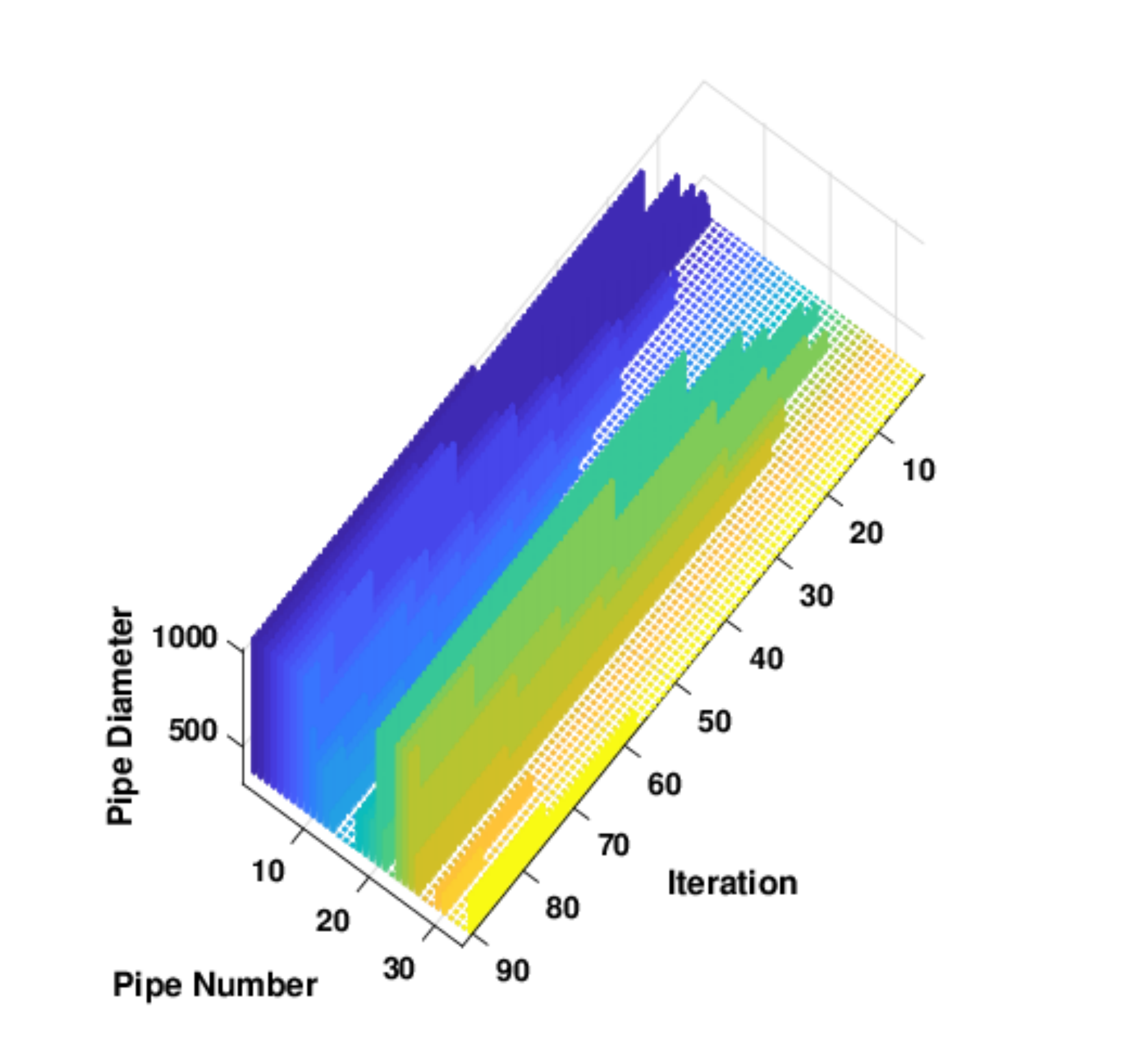}}

\caption{(a) A 3D bar chart of the greedy search performance when all pipes diameters (NYTP) are equal to Zero.(b) for optimizing the Hanoi network when all pipes diameters are initialized to 304.8(mm), maximum number of evaluation is 3094 and the proposed layout cost=\$6,312,405. }%
\label{fig:greedy_search_NYTP}%
\end{figure*}
%------------------------------------
The idea of the hybrid framework third part is a Downward Greedy Search ($GS_D$)(Algorithm \ref{alg:DGS}). The main purpose of the  $GS_D$ is smoothing the pipe cost with respect to the constraints by decreasing the diameter of the pipes one by one. In other words, $GS_D$ is looking for improvements that give us the least reduction in pressure violations for the most significant reduction in the pipe cost. The purpose is maximizing according to the Equation \ref{EQ:DGS}.
%----------------------------------
\begin{equation}
\begin{split} \label{EQ:DGS}
Argmax \to  f(\Theta)=(\frac{\sum_{j=1}^N\triangle PipeCost_{j}}{ \sum_{i=1}^{M}\triangle PV_{i}}) 
\\
Subject-to: \triangle PV_{i} \ge 0 ,\triangle PipeCost_{i}\ge 0
 \end{split}
\end{equation}
%------------------------------------------
Practically, $GS_D$ 
% performance is competitive compared with the combination of Breadth-First search and  Best-First search in terms of computational costs. It
is so fast and effective to attain the near best cheap solution. 
%----- Downward Greedy Search-----------
 
 \begin{algorithm}
 \caption{The Downward Greedy Search (Minimizing the network cost )}\label{alg:DGS}
 \begin{algorithmic}[1]
 \Procedure{ The Downward Greedy Search ($GS_D$)}{}\\
  \textbf{Initialization}
  \State $\mathit{Initialize}~\tau$ \Comment{initialize Pressure head constraint}
  \State $NetPipe_{iter}$=Pipe-Network \Comment{Read Network data }
  \State $Nodal_{Pressure}$=$\mathit{Eval(NetPipe_{iter})}$ \Comment{Evaluate Network by EPANET }
  \State $N=Size(NetPipe_{iter})~and~M=Size(Nodal_{Pressure})$
  \State $Sum_{PV}=\sum_{i=1}^{M} (|\tau-Nodal_{Pressure}^{i}|)~ \forall~ 
   Nodal_{Pressure}^{i} <\tau$ \Comment{Calculate sum of the nodal pressure violation}
  \While{$Sum_{PV}\ge 0$}
    \While{$i\le N$}
  \State Decrease the $i^{th}$Pipe diameter of the $Netpipe_{iter}$ based on the possible diameters
\State $Nodal_{Pressure}$=$\mathit{Eval(Netpipe_{iter})}$ 
 \State $\mathit{Update}~Sum_{PV}$
 \If{$Sum_{PV}=0$}
 \State Add the $NetPipe_{iter}$ to feasible solution set
  \State $Improvement_{rate}^{i}=\frac{ \sum_{i=1}^{N}\triangle PipeCost_{i}}{\sum_{i=1}^M\triangle PV_{i}}$
  \EndIf
 \EndWhile
 \State $NetPipe_{iter}$=Max($Improvement_{rate}$) \Comment{Select the best feasible solution}
    \EndWhile
    \State return $NetPipe_{iter}$
 \EndProcedure
 \end{algorithmic}
 \end{algorithm}

 %---------------------------------------------------------------------
 \begin{algorithm}
 \caption{The CM$A_{ES}$-G$S_U$-G$S_D$}\label{alg:CMAGS}
 \begin{algorithmic}[1]
 \Procedure{ The CM$A_{ES}$-G$S_U$-G$S_D$}{}\\
  \textbf{Initialization}
  \State Initialize the CMA-ES Parameters
  \State $\mathit{\lambda}$ \Comment{Offspring population size}
  \State $\mathit{\mu}$   \Comment{Parent population size ($floor(\lambda/2)$)}
  \State $\mathit{\sigma_{start}}$ \Comment{Initial standard deviation($0.5\times (UB-LB)$)}
  \State $\mathit{c_{c}}$ \Comment{Covariance learning rate}
%   \State $D, \lambda, \mu, \sigma_{strat}, c_{c}$, $Pop_{Size}(NP),\mu$, $w_i \forall 1,...,\mu,$, $\mu_{eff},c_{\sigma},c_1,c_{\mu}, d_{\sigma}$ $P_{\sigma}=P_{c}$=0, C=1, $B=I_n$
%   \State $Pop_{1}= unifrnd(VarMin,VarMax,[Pop_{size} 1])$
    \While{$Function Tolerance \le \xi$}
\State Update Evolutionary Parameters
 \State $m,P_{\sigma}, P_{c}, C,\sigma$ are updated.Computing B,D through eigen decomposition of C.
\State $C^{(g+1)}=(1-c_c)C^g+\frac{c_c}{\mu_c}P_c^{(g+1)}P_c^{(g+1)^T}+c_c(1-\frac{1}{\mu_c})\times \sum_{i=1}^x w_i (\frac{X_{1:\lambda}^{(g+1)}-\mu^g}{\sigma^{(g)}})(\frac{X_{1:\lambda}^{(g+1)}-\mu^g}{\sigma^{(g)}})^T$
\State Update the mutation step size $\sigma^g$
\State $\sigma_{g+1}=\sigma_g\times exp(\frac{c_\sigma}{d_\sigma}(\frac{\parallel P_\sigma \parallel}{EN(0,I)}-1))$ 
\State Generate sample population for next generation (g+1)
\State 
$x_k^{(g+1)}\sim \mathcal{N}(m^{(g)},(\sigma^{(g)})^2C^{(g)}) \forall k=1,...,\lambda$
\If{($x_k^{(g+1)}$ is not feasible)}
 \State  Impose the penalty
 \EndIf
\State Update the mean for next generation (g+1)
\State $m^{(g+1)}=\sum _{i=1}^\mu w_ix_{i:\lambda}^{(g+1)}$
\State Update best ever solution ($\mathit{Best_{Solution}}$)
\State $\mathit{Possible_{Solution}}$=round($\mathit{Best_{Solution}}$) \Comment{Convert continuous design to possible pipe size}
\State $\mathit{Sum_{PV}}$=Sum-Violation$(\mathit{Possible_{Solution})}$ \Comment{Compute sum of the nodal pressure violation}
 \If{($ \mathit{f(Possible_{Solution})} < \phi ~\&~ Sum_{PV}>0 $))}
 \State  Apply the Upward Greedy Search ($GS_U$) \Comment{Fix the violation of nodal pressure heads}
 \EndIf
     \EndWhile\
     
 \State  Apply the Downward Greedy Search($GS_D$)
 \EndProcedure
 \end{algorithmic}
 \end{algorithm}
%-----------------------------------------------
\section{WDS design formulation and constraints} \label{sec:WDS-details}
In this research, the main optimization aim is minimizing the cost of the pipes of the network concerning the constraints. The WDSs optimization problem is a combinatorial optimization which can be defined as ascertaining the best mixture of the element dimensions and settings such as the size of pipe diameters (as the decision variables), pump types and so on that supplies the least cost for the yielded network design. Although pipe layout and its connectivity, the pattern of the nodal demand, and minimum nodal head conditions should be fulfilled. The optimization problem is declared mathematically as :
%----------------------------------------
\begin{equation}
%\begin{split}
\begin{aligned}
\label{EQ:OBF}
Argmin \to  C_{pipe}(\Theta)&=
\sum_{i=1}^N(c_i\times D_i) \times L_i 
\end{aligned}
\end{equation}

$Subject-to:
\\
(1): H_j \ge H_j^{min}, \quad \quad \forall \quad j=1,...,M
\\
(2): D_k \in \{D\},\quad \quad \forall \quad k\in N$\\
%-------------------------------------------
Where $C_{pipe}$ is the primary fitness function including the layout cost and also  $L_i$ and $D_i$ are the $i^{th}$pipe length and diameter respectively; $c_i$ can be the cost per unit length of pipe in the network. The above cost function is constrained to the following.

Firstly, it is the minimum nodal pressure constraint that should be imposed for all nodes of the network. The $H_j$ points out the pressure level in the $j^{th}$ node, and the minimum needed $j^{th}$ nodal pressure based on the demand pattern is shown by $H_j^{min}$. If this constraint is not satisfied the sum of nodal head pressure violation will be computed (Equation \ref{sum_violation}). 
%-----------------------------------
\begin{fleqn}
\begin{align} \label{sum_violation}
Sum_{PV}=
\begin{cases}
0& if (H_j \ge H_j^{min})   \\
\sum_{j=1}^M(H_j^{min}-H_j)&     otherwise 
\end{cases}
\end{align}
\end{fleqn}
%---------------------------------------
The second constraint is the possibility of the discrete pipe sizes, which are defined commercially. Thus, if the diameter of pipes does not include from the discrete sizes set, the fitness function must be penalized by the Equation \ref{sum_D_violation}. Where $D_k$ mentions the diameter of pipe $i$ that is chosen from a deterministic set (D). 

%---------------------------------------
\begin{fleqn}
\begin{align} \label{sum_D_violation}
Sum_{DV}=\sum_{i=1}^N
\begin{cases}
0& D_i = D_a||D_b \\
1&   D_i = (D_a+D_b)/2   \\
\frac{D_i-a}{((D_a+D_b)/2)-D_a}& D_i < (D_a+D_b)/2 \\
\frac{D_b-D_i}{D_b-((D_a+D_b)/2)}& D_i > (D_a+D_b)/2
\end{cases}
\end{align}
\end{fleqn}
%----------------------------------------

Where both $D_a$ and $D_b$ are sequential possible pipe diameters ($D_a < D_b$) and $D_i$ is the size of pipe i which can be possible diameter or continuous.  When the sum of the violation is not equal to zero, the constrained model is transformed into an unconstrained one by inserting the sum of constraint violations value to the fitness function as the penalties. While all computations of the hydraulic simulation are done via EPANET 2.0. Consequently, the total cost is acknowledged as the sum of the pipe cost, a penalty cost of pressure violation and the violation of discrete pipe diameters represented as:

%--------------------------------------------
\begin{equation}
%\begin{split}
\begin{aligned}
\label{EQ:cost}
C_{PV}(\Theta)&=(P_f\times Sum_{PV})  \\
C_{DV}(\Theta)&=(P_D\times Sum_{DV}) \\
Minimize \to  C_{t}(\Theta)&= C_{pipe}(\Theta)+C_{PV}(\Theta)+ C_{DV}(\Theta)
 \end{aligned}
\end{equation}
where $C_t$ is the total cost of the penalized or non-penalized fitness function cost, and also both $P_f$ and $P_D$ are the penalty factors. For instance, Algorithm \ref{alg:penalty} shows how the penalty of the continuous pipe diameters is imposed for the NYTP and NYTP2.

%----------------------------------------------
%-----------------------------------------
\begin{algorithm}
\caption{Handling constraint violations for continuous pipe diameters of NYTP}\label{alg:penalty}
 \begin{algorithmic}
 \Procedure{ The Penalty function}{}
 \If{$rem(D_i,12)\ne0$}
\If{$D_i > 36$}\\
$Sum_{DV}=\sum_{i=1}^N$
 $\begin{cases}
 0& D_i = D_a||D_b \\
1&   D_i = (D_a+D_b)/2   \\
\frac{D_i-a}{((D_a+D_b)/2)-D_a}& D_i < (D_a+D_b)/2 \\
\frac{D_b-D_i}{D_b-((D_a+D_b)/2)}& D_i > (D_a+D_b)/2
\end{cases}$
\Else\\
$Sum_{DV}=\sum_{i=1}^N$
 $\begin{cases}
 0 & D_i=0| D_i=36\\
\frac{D_i}{18}*3 &D_i < 18\\
3 & D_i = 18\\
\frac{36-D_i}{18}*3 &D_i > 18
\end{cases}$
 \EndIf
  \EndIf
   \EndProcedure
 \end{algorithmic}
 \end{algorithm}
%------------------------------------------------------
%--------------------------------------------------------
\section{Case Studies Results and Discussions} \label{sec:results}
For evaluating the proposed hybrid framework effectiveness, five well-known WDSs case studies have been applied including the New York Tunnels Problem (NYTP) \cite{dandy1996improved}, the Doubled New York Tunnels Problem (NYTP2) \cite{zecchin2005parametric}, 50NYTP \cite{zheng2011optimal}, the Hanoi Problem (HP) \cite{fujiwara1990two} moreover, one large-scale network called the Balerma Network (BN) \cite{reca2006genetic}. The details of the case studies can be seen in Table \ref{details_networks}. 
%---------------------------------
\begin{table}
\caption{The characteristics of Case Studies Summery }\label{details_networks}
\centering
%\scalebox{0.85}{
\begin{tabular}{l|p{3cm}|p{2.3cm}|p{2.5cm}|p{3cm}}
\hlineB{4}
\textbf{WDS} &\textbf{Decision Variables} & \textbf{No. of options} &\textbf{No. of Nodes}& \textbf{Search Space Size}  \\  \addlinespace[0.1cm]\hlineB{4}
  NYTP &\centering 21    &\centering16   &\centering20&$1.934\times 10^{25}$   \\ \hline
  HP   &\centering34    &\centering6    &\centering32&$2.865\times 10^{26}$   \\ \hline
  NYTP2&\centering42    &\centering16   &\centering20&$3.741\times 10^{50}$    \\ \hline
  BN   &\centering454   &\centering10   &\centering447&$ 10^{454}$   \\ \hline
  NYTP50&\centering1050    &\centering16   &\centering1000&$2.12\times 10^{1264}$    \\ \hlineB{4}
  \end{tabular}
%  }
\label{Summery_Character}
\end{table}
%-----------------------------------
\subsection{Case Study 1: New York Tunnel problem (NYTP)}

The NYTP layout is a fundamental benchmark of the water distribution system problem which is inspired by the real New York water network.  
The number of existing tunnels is 21 with 20 nodes supported by a fixed-head reservoir. The detailed information of NYTP  provides by Dandy et al. \cite{dandy1996improved} such as the cost of pipes, nodal demand pattern and the head constraints.  The principal purpose is minimizing the total pipe cost of the new installed parallel pipes, while the existing pipes are accompanied. Meanwhile,  the constraints should be handled (minimum nodal pressure head).

In NYTP,  pipes diameter size can be allocated among 15 actual different sizes plus a zero size that means in this tunnel; there is not any new pipe. Therefore, the search space size is $16^{21}$.  However, in this research, a continuous problem space is considered too. The benchmarks have been evaluated by different aspects of search space like as continuous, discrete (interval=1 inch) and possible (commercialized) layout. With regard to assessing the ability of the proposed CM$A_{ES}$-G$S_U$-G$S_D$ algorithm to achieve a great balance between exploration and exploitation in decision space a range of population sizes, as specified by the maximum iteration number, are considered for each case study network such as $\lambda=$10, 20, 50, 100, 200 and 400. Since the CMA-ES is a self-adaptive method, all control parameters have been adjusted during the optimization process except $\sigma$. The  $\sigma$ is initialized by half of the decision variables length. Thus, the CMA-ES is started by a considerable ability of exploration.

In the first step of the proposed hybrid framework, the CMA-ES efficiency is evaluated by three kinds of decision variables: continuous, discrete and possible. Where the continuous pipe sizes are used, the only nodal pressure head constraint should be satisfied, so the penalty factor ($PF$) for NYTP is $10^7$. Moreover, a severe penalty factor is imposed too, which can be seen in the Algorithm \ref{alg:penaltyPressure}. According to the achieved results, the performance of the severe  $PF$ is not competitive. 

The best configuration of the NYTP cost which is obtained by continuous CMA-ES is considerable at \$38.00 million that is a feasible solution based on the nodal pressure head constraint by different population sizes (The best-known NYTP cost is \$38.64 million \cite{eiger1994optimal}). Thus, CMA-ES can overcome all previous methods in continuous search space. The proposed new feasible continuous designs of NYTP are declared in the Table \ref{Min_Feasible_continuous} and also a comprehensive review comparison of the best-founded NYTP layouts that some of them are infeasible after evaluating the nodal head constraints are listed in Table \ref{table_review_NYTP}.

\begin{figure}
 \centering
\includegraphics[ width=0.6\textwidth]
{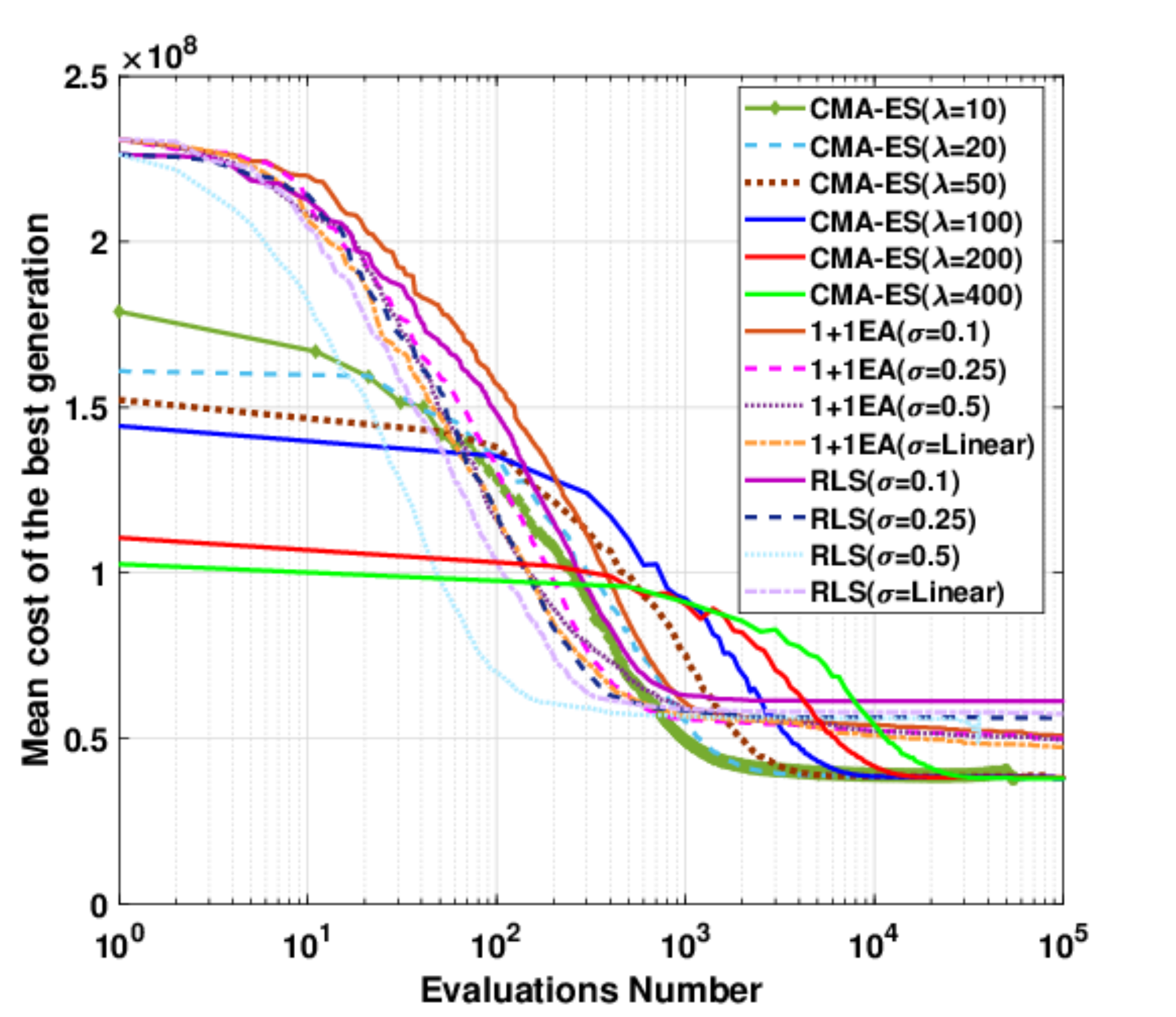}
 \caption{ The comparison of the average convergence rate of CMA-ES ,RLS and 1+1EA for the {continuous NYTP} design where  $\lambda=10, 20, 50, 100, 200, 400$ and $\sigma=0.1, 0.25, 0.5, Linear.$ } \label{fig:NYTP_convergence}
 \end{figure}

In EAs, the convergence rate is another significant evolutionary parameter to realize how fast EAs converge to the optimal solutions per generation.  In this way, Figure \ref{fig:NYTP_convergence} represents the average convergence rate of the proposed methods by 30 independent runs. We can see the CMA-ES with small population sizes has converged faster compared with the big population sizes. However, mostly it faces with the premature convergence situation and falling into a local optimum.  
%----------------------------------------------
\begin{algorithm}
\caption{Penalizing the pressure violations }\label{alg:penaltyPressure}
 \begin{algorithmic}
 \Procedure{ The Penalty function for handling the pressure violations(PV)}{}
 \If{$\sum_{i=1}^{i=N}PV_(i)>0$}\\
$Penalty_{(PV)}=10^5+(10^4*\sum_{i=1}^{i=N}PV_{(i)})^4$
 \EndIf
    \EndProcedure
 \end{algorithmic}
 \end{algorithm}
%----------------------------------------
%--Feasible Continuous, Discrete Layouts-----
 \begin{table*}
 \centering
\caption{The Achieved best feasible continuous (C) and Discrete (D) NYTP design by CMA-ES. Next, the best solutions are rounded to possible pipe diameters ($r$), but rounded layouts have a negligible   nodal pressure violation ($\sim$).}
\scalebox{0.55}{
\begin{tabular}{p{1.5cm}|p{3cm}|p{3cm}|p{3cm}|p{3cm}|p{2.6cm}|p{2.8cm}|p{2.8cm}|p{2.5cm}}
\hlineB{4}
\textbf{Pipe} &\textbf{CMA-ES ($C,\lambda$=10)} & \textbf{CMA-ES ($C,\lambda$=20)}&\textbf{CMA-ES ($C,\lambda$=50)}&\textbf{CMA-ES ($C,\lambda$=100)} & \textbf{CMA-ES ($D,\lambda$=10)} & \textbf{CMA-ES ($D,\lambda$=20)}&\textbf{CMA-ES ($D,\lambda$=50)}&\textbf{CMA-ES ($D,\lambda$=100)}   \\ \hline \hlineB{4}

  1  &0&0&0&0 &0 &0 &0 &0   \\ \hline
  2  &0&0&0&0&0 &0 &0 &0 \\ \hline
  3  &0&0&0&0&0 &0 &0 &0  \\ \hline 
  4  &0&0&0&0&0 &0 &0 &0 \\ \hline
  5  &0&0&0&0&0 &0 &0 &0   \\ \hline
  6  &0&0&0&0&0 &0 &0 &0    \\ \hline
  7 
  &119.06 $\to 120_{r}$
  &118.99 $\to 120_{r}$
  &118.98 $\to 120_{r}$
  &118.99 $\to 120_{r}$
  &0
  &118.12 $\to 120_r$
  &117.19 $\to 120_r$
  &122.09 $\to 120_r$
    \\ \hline
  8  &0&0&0&0&0 &0 &0 &0   \\ \hline 
  9  &0&0&0&0&5.084 $\to 0.0_r$ &0 &0 &0  \\ \hline
  10 &0&0&0&0&0 &0 &0 &0   \\ \hline
  11 &0&0&0&0&0 &0 &0 &0    \\ \hline
  12 &0&0&0&0&6.895 $\to 0.0_r$&0&0 &0  \\ \hline
  
  13 
  &0
  &0
  &0
  &0
  &0
  &0
  &0
  &0
    \\ \hline 
  
  14 
  &0
  &0
  &0
  &0
  &0
  &0
  &0
  &0
    \\ \hline
  
  15
  &0
  &0
  &0
  &0
  &120.01 $\to 120_r$
  &0
  &0
  &0
  \\ \hline
  
  16 
  &99.95 $\to 96_{r}$
  &99.97 $\to 96_{r}$
  &99.98 $\to 96_{r}$
  &99.97 $\to 96_{r}$
  &82.01 $\to 84_r$ 
  &100.14 $\to 108.0_r$ 
  &100.76  $\to 96.0_r$ 
  &99.06 $\to 96_{r}$
    \\ \hline
  
  17 
  &99.29 $\to 96_{r}$
  &99.28 $\to 96_{r}$
  &99.28 $\to 96_{r}$
  &99.28 $\to 96_{r}$
  &101.31  $\to 96.0_r$
  &99.97  $\to 96.0_r$ 
  &98.71   $\to 96.0_r$ 
  &99.31 $\to 96_{r}$
  \\ \hline
  
  18 
  &79.04 $\to 84_{r}$
  &79.08 $\to 84_{r}$
  &79.09 $\to 84_{r}$
  &79.08 $\to 84_{r}$
  &74.32 $\to 72.0_r$
  &78.86  $\to 84.0_r$
  &80.04  $\to 84.0_r$
  &79.01 $\to 84_{r}$
  \\ \hline 
  
  19 
  &75.04 $\to 72_{r}$
  &75.07 $\to 72_{r}$
  &75.05$\to 72_{r}$
  &75.06  $\to 72_{r}$
  &68.14 $\to 72.0_r$
  &81.63  $\to 84.0_r$ 
  &75.96  $\to 72.0_r$
  &76.23 $\to 72_{r}$
   \\ \hline
  
  20 &0&0&0&0&0&0&0&0 \\ \hline
  
  21 
  &70.61 $\to 72_{r}$
  &70.60 $\to 72_{r}$
  &70.61 $\to 72_{r}$
  &70.61 $\to 72_{r}$
  &73.65 $\to 72.0_r$
  &69.03  $\to 72.0_r$
  &70.74  $\to 72.0_r$
  &70.13  $\to 72_{r}$
  \\ \hline \hlineB{4}
  
  Total Cost
  &\textbf{38.00} $\to 37.63_{r}^{\sim}$ 
  &\textbf{38.00} $\to 37.63_{r}^{\sim}$ 
  &\textbf{38.00} $\to 37.63_{r}^{\sim}$ 
  &\textbf{38.00} $\to 37.63_{r}^{\sim}$ 
  &\textbf{38.57} $\to 37.69_r^\sim$
  &\textbf{38.26} $\to 39.61_r^\sim$
  &\textbf{38.09} $\to 37.63_r^\sim$
  &\textbf{38.05}
  \\ \hline \hlineB{4}
   $16^{th}$
   & 1.42e-07 $\to$ -0.042
   & 5.35e-07 $\to$ -0.042
   & 2.49e-07 $\to$ -0.042
   & 1.15e-07 $\to$ -0.042
   & 0.08 $\to$ -0.255
   & 0.16 $\to$ 0.68
   & 0.073 $\to$ -0.042
   & 0.025
   \\ 
    $17^{th}$
    &1.13e-09 $\to$ -0.046
    &5.55e-10 $\to$ -0.046
    &1.04e-08 $\to$ -0.046
    &2.63e-08 $\to$ -0.046
    &0.003 $\to$ 0.047
    &8.20e-04 $\to$ 0.085
    &0.002 $\to$ -0.046
    &0.002
   \\ 
    $19^{th}$
   Nodal Pressure
   & 1.47e-06 $\to$ -0.044
   &  3.03e-07 $\to$ -0.044
   &  3.04e-07 $\to$ -0.044
   &  6.02e-07 $\to$ -0.044
  &  0.036 $\to$ -1.38
  & 0.09  $\to$ -0.0455 
  & 0.015$\to$  -0.044
  & 0.008 
   \\\hlineB{4}
\end{tabular}
}

\label{Min_Feasible_continuous}
\end{table*}
%-------------------------------------
%---Possible NYTP (rounded)----------------
  \begin{table*}
 \centering
\caption{A comparison results of CMA-ES and CM$A_{ES}-GS_U$ (Hybrid method of CMA-ES and Upward Greedy search) for {NYTP}, All proposed networks by CM$A_{ES}-GS_U$ are feasible in terms of both pipe diameter and nodal pressure.  }
\scalebox{0.9}{
\resizebox{\textwidth}{!}{\begin{tabular}{p{1.8cm}?p{1.8cm}|p{2.3cm}?p{1.8cm}|p{2.3cm}?p{1.8cm}|p{2.3cm}?p{1.8cm}|p{2.3cm}?p{1.8cm}|p{2.3cm}?p{1.8cm}|p{2.3cm}}
\hlineB{4}

\textbf{Pipe} &\textbf{CMA-ES} & \textbf{CM$A_{ES}-GS_U$}&\textbf{CMA-ES} &\textbf{CM$A_{ES}-GS_U$}&\textbf{CMA-ES}  &\textbf{CM$A_{ES}-GS_U$}  &\textbf{CMA-ES} &\textbf{CM$A_{ES}-GS_U$}  &\textbf{CMA-ES} &\textbf{CM$A_{ES}-GS_U$} &\textbf{CMA-ES }&\textbf{CM$A_{ES}-GS_U$}   \\ \hline \hlineB{4}
  1  &0  &0  &0  &0  &0  &0  &0  &0  &0  &0  &0  &0  \\ \hline
  2  &0  &0  &0  &0  &0  &0  &0  &0  &0  &0  &0  &0  \\ \hline
  3  &0  &0  &0  &0  &0  &0  &0  &0  &0  &0  &0$\to$  &\textbf{36} \\ \hline 
  4  &0  &0  &0  &0  &0  &0  &0  &0  &0  &0  &0  &0  \\ \hline
  5  &0  &0  &0  &0  &0  &0  &0  &0  &0  &0  &0  &0  \\ \hline
  6  &0  &0  &0  &0  &0  &0  &0  &0  &0  &0  &0  &0  \\ \hline
  7  &0  &0  &120 $\to$&\textbf{144}&144&144&0  &0  &120 $\to$&\textbf{144}&0  &0  \\ \hline
  8  &0  &0  &0  &0  &0  &0  &0  &0  &0  &0  &0  &0  \\ \hline 
  9  &0  &0  &0  &0  &0  &0  &0  &0  &0  &0  &0  &0  \\ \hline
  10 &0  &0  &0  &0  &0  &0  &0  &0  &0  &0  &0  &0  \\ \hline
  11 &0  &0  &0  &0  &0  &0  &0  &0  &0  &0  &0  &0  \\ \hline
  12 &0  &0  &0  &0  &0  &0  &0  &0  &0  &0  &0  &0  \\ \hline
  13 &0  &0  &0  &0  &0  &0  &0  &0  &0  &0  &0  &0  \\ \hline 
  14 &0  &0  &0  &0  &0  &0  &0  &0  &0  &0  &0$\to$  &\textbf{60} \\ \hline
  15 &120&120&0  &0  &0  &0  &132$\to$&\textbf{144}&0  &0  &0  &0  \\ \hline
  16 &84 &84 &96 &96 &84$\to$ &\textbf{96} &72 &72 &96 &96 &108 $\to$&\textbf{132}\\ \hline
  17 &96 &96 &96 &96 &96 &96 &96 &96 &96 &96 &96 &96 \\ \hline
  18 &72 $\to$ &\textbf{84} &84 &84 &84 &84 &72$\to$ &\textbf{84} &84 &84 &84 &96 \\ \hline 
  19 &72 &72 &72 &72 &72 &72 &60$\to$ &\textbf{72} &84 &84 &60 &60 \\ \hline
  20 &0  &0  &0  &0  &0  &0  &0  &0  &0  &0  &0  &0  \\ \hline
  21 &72 &72 &72 &72 &72 &72 &72 &72 &72 &72 &84 &84 \\  \hlineB{4}
  Total Pipe Cost (\$M)&\textbf{37.70} & {\textbf{38.81}} &\textbf{37.63} &{\textbf{38.64}}  &\textbf{37.37} &{\textbf{38.64}}  &\textbf{36.63}  &\textbf{39.22}  &\textbf{38.30} &\textbf{39.31} & \textbf{35.52} &\textbf{43.81} \\ \hlineB{4}
     $16_{th}$,
     &0.255
     &0.255
     &0.048
     &0.033
     &0.033
     &0.033
     &-0.894 
     &0.513
     &0.676
     &0.750
     &0.033
     &0.425
     \\
     $17_{th}$
      &$0.0476$
     &$-0.0419$
     &$-0.0463$
     &$0.0296$
     &$-0.1664$
     &$0.0296$
     &$-0.1266$ 
     &$0.0094$
     &$-0.0427$
     &$-0.3011$
     &$-0.0257$
     &$0.0022$
     \\
     $19_{th}$ Pressure node&  $-1.3785$&    $0.3372$&   $-0.0432$&   $0.0234$  &   $0.0234$ &   $0.0234$ &$-1.2365$  &   $0.6129$ &   $-0.0455$ &   $0.0215$  &  $-0.3635$ &   $0.8309$  \\ \hlineB{4}
\end{tabular}}
}
\label{Min_possible_NYTP}
\end{table*}
 %-------------------------------------------
%---------------------------------------
 \begin{figure}
 \centering
\includegraphics[ width=0.7\textwidth]
{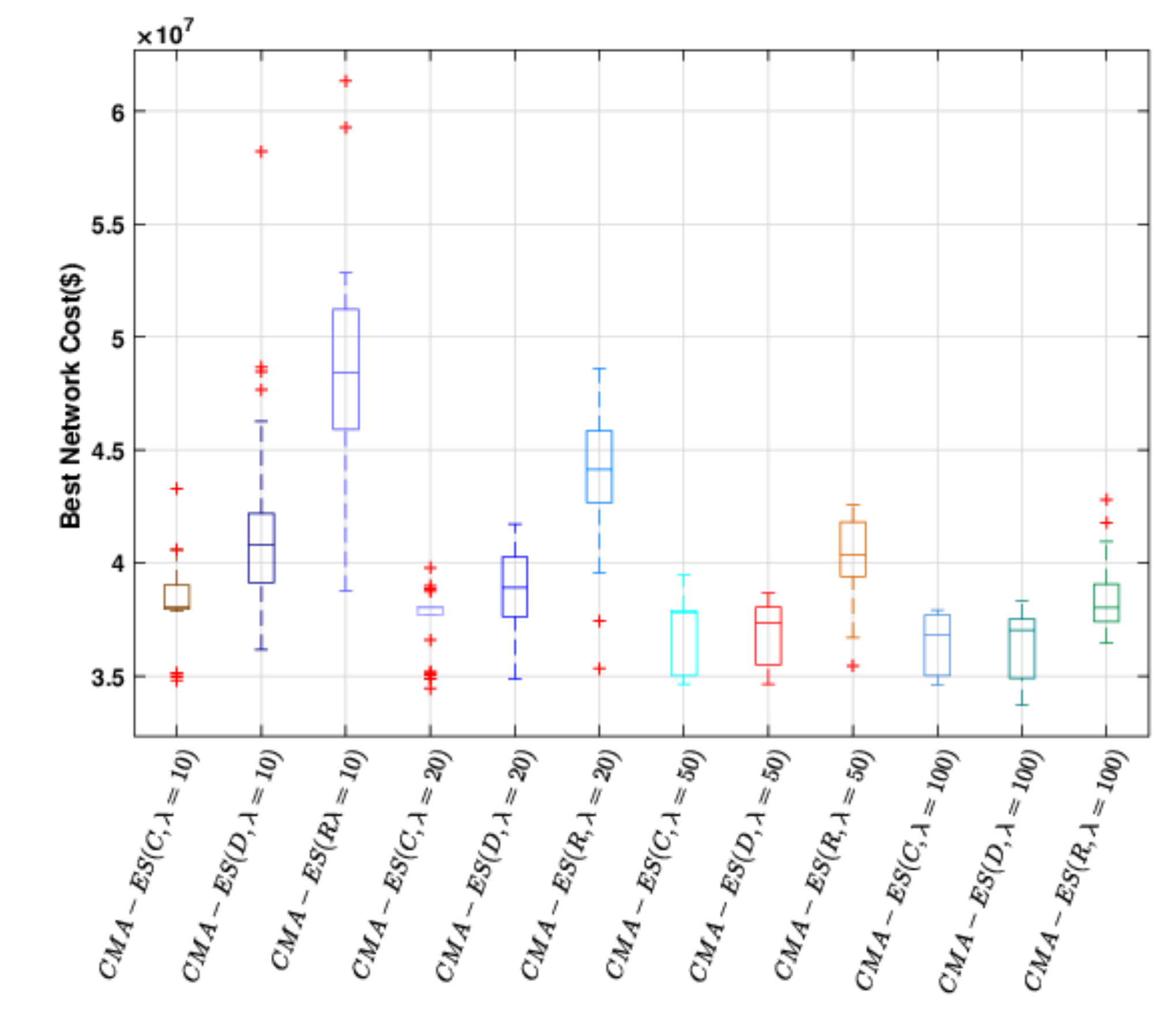}
 \caption{ A comparison of the three different strategies for Optimizing the New York Tunnel problem by different CMA-ES population sizes ($\lambda=10, 20, 50, 100$). C= Continuous, D= Discrete and R= Rounded (Possible)  , the best solution per experiment, 30 independent runs. According to the results, big population sizes can be a better choice like 50 or 100, and also continuous pipe diameter strategy beats both discrete and rounded pipe size methods. However, the sum of the nodal head violation is not zero in some of the cheap designs.}   \label{fig:comp_all_NYTP}
 \end{figure}
 %-------------------------------------------
%---------------------------------------------
\begin{figure}
\centering
\includegraphics[ width=0.55\textwidth]
{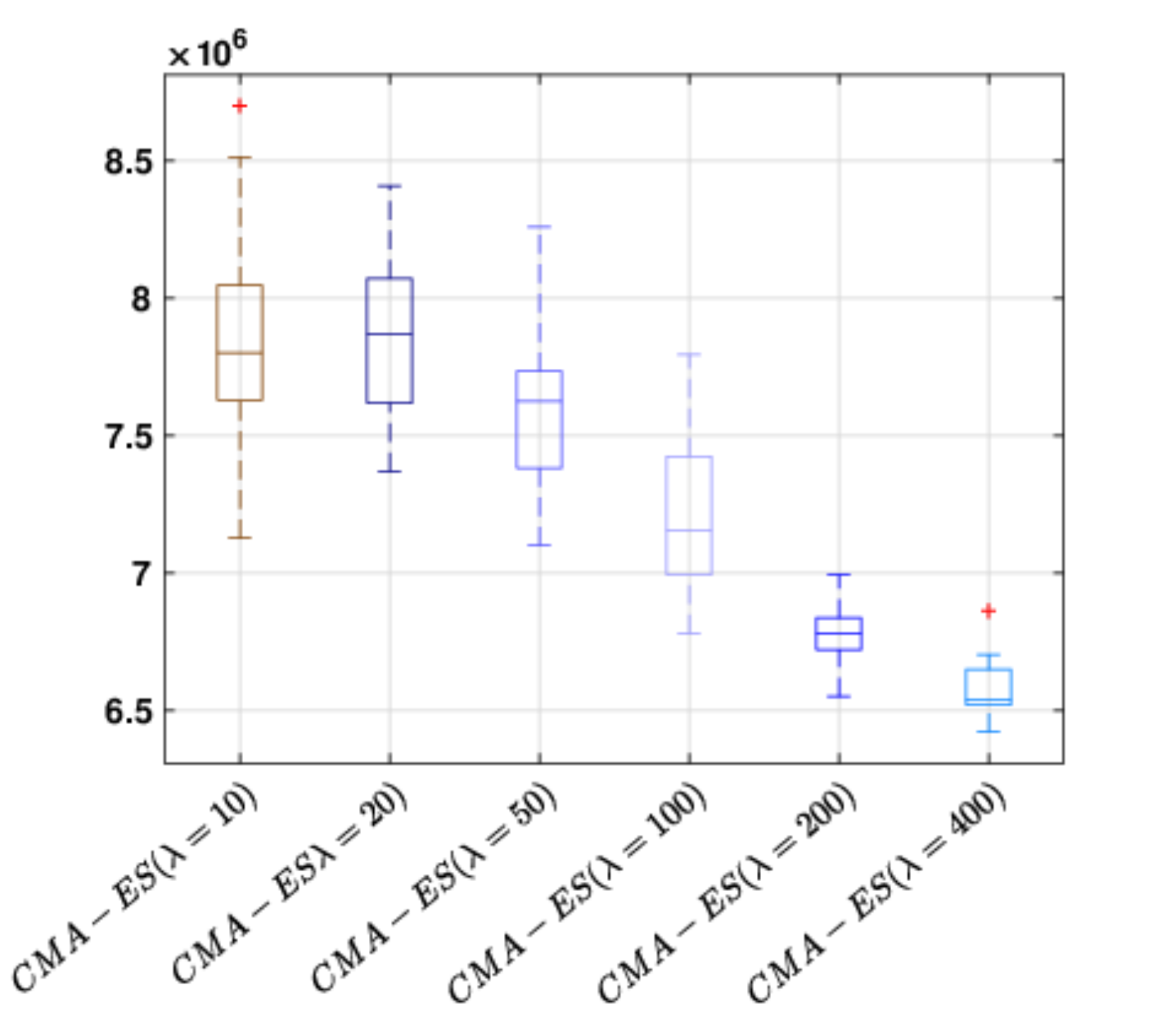}
\caption{ The impact of different population sizes of CMA-ES with the discrete penalty cost in the {Hanoi} network.} \label{fig:Dis_Hanoi}
\end{figure}
%---------------------------------------------------------
%---------------------------------------------
\begin{figure}
\centering
\includegraphics[ width=\textwidth]
{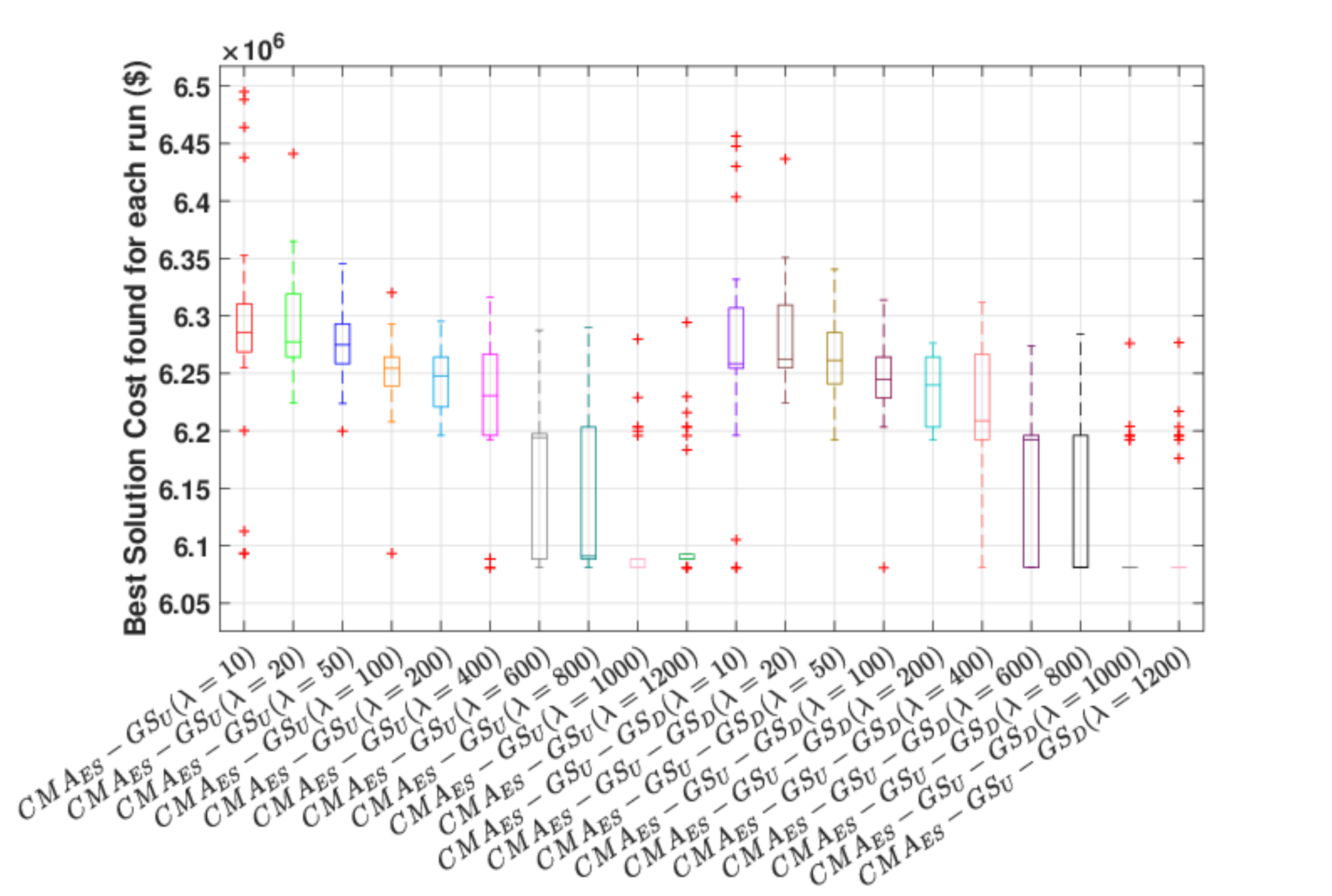}
\caption{ The impact of different population sizes on the CMA$_{ES}-GS_U$ and   CMA$_{ES}-GS_U-GS_D$ in the {Hanoi} network.} \label{fig:Hanoi-boxplot}
\end{figure}
%--------------------------------
%--------------------------------------
 %------------------------------------
 Besides, between two other strategies, the discrete pipe size scenario when the interval is 1 inch has better performance than the rounded (possible) pipe size. Figure \ref{fig:comp_all_NYTP} shows the comparison of the three strategies and the impact of large population sizes. As the CMA-ES is able to discover very cheap designs when the pipe sizes are discrete or rounded (possible), but they are not feasible in terms of nodal head constraints, the Upward Greedy search is applied to fixing up the designs.

For saving the computational budgets of the CMA-ES, the termination criterion is configured as a function tolerance value ($\xi$) at $1/10^5\times C_{t}(\Theta)$. The $GS_U$ will be begun if the network cost of the rounded pipes is less than $\phi$ (well-known or estimated network cost). The efficiency of $GS_U$ is independently tested by the initialization of zero pipe sizes. The $GS_U$ is a super fast search method and can find an NYTP layout at  \$42.36 million just by  714 evaluations number (Figure \ref{fig:greedy_search_NYTP}). Substantially, the $GS_U$ is able to fix up the violation of nodal pressure head by increasing some of the pipe sizes all the time. This combination of CMA-ES and $GS_U$ leads to a robust and powerful framework that finding the optimal feasible solutions are guaranteed. The best results have occurred when the CM$A_{ES}$-G$S_U$ population size is 400 that the percentage of reaching the best well-known NYTP layout is 100\%. Table \ref{Min_possible_NYTP} illustrates how CM$A_{ES}$-G$S_U$  explores and finds the feasible NYTP designs with possible pipe diameter sizes compared with the CMA-ES.

The best average evaluations number get the best solution for the NYTP  is 5400 for CMA-ES ($\lambda=20$) and 5500 for CM$A_{ES}$-G$S_U$($\lambda=20$). Both rates are less than those of other reported methods.
Furthermore, where the population sizes are 10, 20, 50 and 100, The maximum admissible evaluations for NYTP, NYTP2 and HP case studies is $10^5$, and for bigger population sizes, those numbers are  $2\times 10^5 $.

\begin{figure}
 \centering
\includegraphics[ width=\textwidth]
{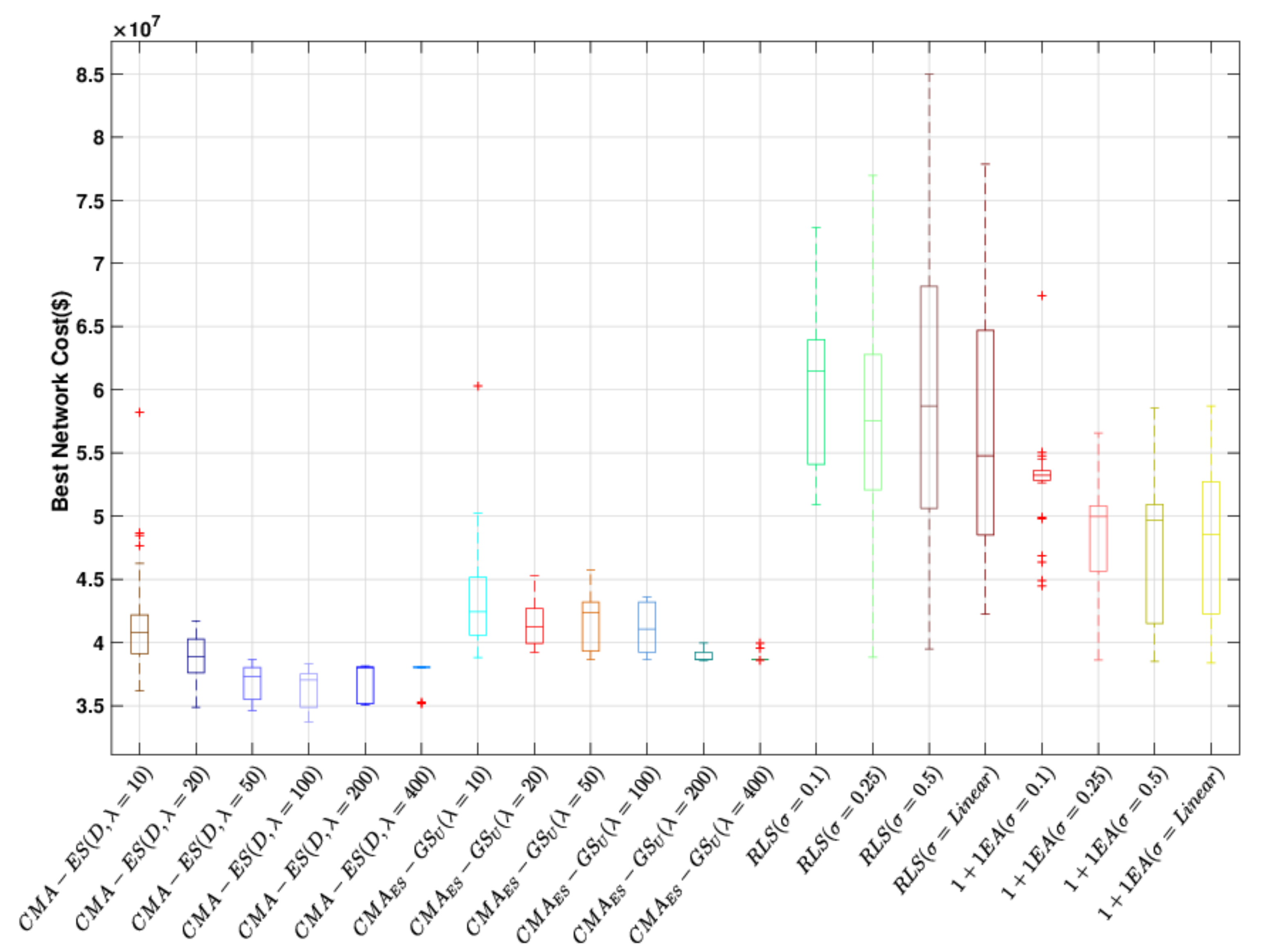}
\caption{Optimizing the {Discrete(interval=1(inch))} New York Tunnel Network  design by different CMA-ES population sizes ($\lambda$=10, 20, 50, 100, 200, 400), CM$A_{ES}-GS_U$, RLS and 1+1EA($\sigma$=0.1,0.25,0.5 $\times$(UB-LB) and Linear) , the best solution per experiment, 30 independent runs, Maximum Evaluation number=$10^5$($\lambda$=10, 20, 50, 100)  and $2\times10^5$($\lambda=200$) as well as $\lambda=400$.{ Remark: all discovered solutions by CM$A_{ES}$-G$S_U$ are feasible in terms of both pipe diameters and pressure head constraints, Best NYTP cost= \$38.64 (Million),but some of the discrete CMA-ES designs are not feasible.}  } \label{fig:comp_discrete}
 \end{figure}

\begin{figure*}
 \centering
\includegraphics[ width=\textwidth]
%{continuous_comparison.eps}
{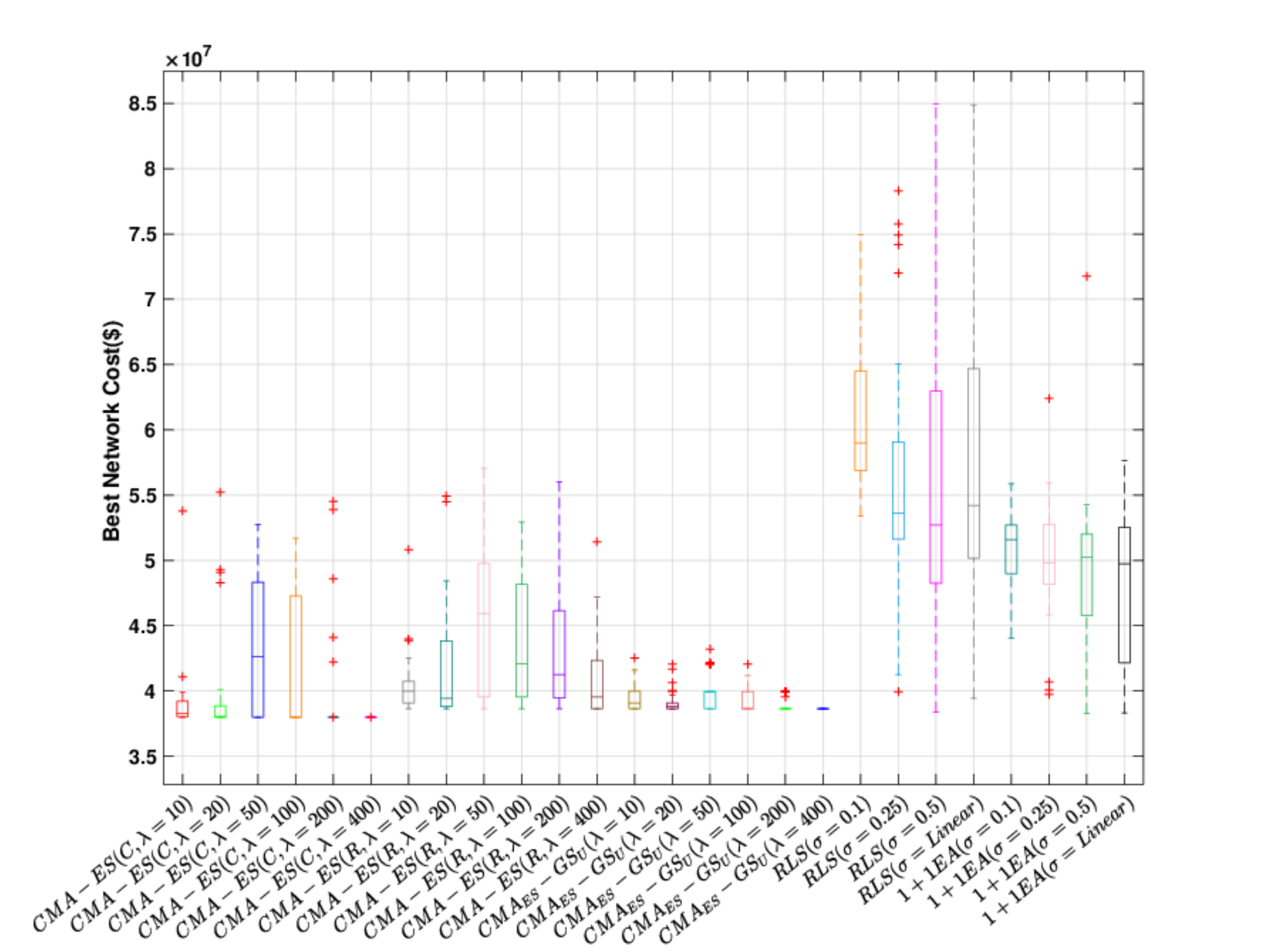}
 \caption{ Optimizing the NYTP {continuous} design by different CMA-ES population sizes ($\lambda$=10, 20, 50, 100, 200, 400), CM$A_{ES}-GS_U$, RLS and 1+1EA ($\sigma$= 0.1, 0.25, 0.5 $\times$ (UB-LB) and Linear) , the best solution per experiment, 30 independent runs, Maximum Evaluation number=$10^5$ ($\lambda$=10, 20, 50, 100), 2$\times10^5$ for $\lambda=200$ as well as $\lambda=400$. C and R show continuous and rounded pipe diameters designs. All solutions are feasible in terms of nodal head pressure. } \label{fig:comp_continuous}
 \end{figure*}

Both fine-tuned simple EAs are assessed to minimize the cost of NYTP by four various mutation step sizes including $\sigma=0.1,0.25,0.5 \times length(decision- variables)$ and a linear mutation step size which is decreased linearly. The best cost of the RLS founded solution is \$39.43 million with a linear $\sigma$ and for the (1+1)EA is \$38.88 million when $\sigma$ is equal to 0.5. Both methods are not able to find the best-known solution of NYTP, and it shows the high complexity of the problem. The summary of the detailed outcomes of NYTP case study can be presented in Table \ref{summary_per_NYTP}. For having a comprehensive analysis of the proposed hybrid framework and both RLS and (1+1)EA effectiveness, The box plot of the NYTP results is drawn with the continuous decision variables (Figure \ref{fig:comp_continuous}) as well as with the discrete (interval=1(inch)) pipe sizes (Figure \ref{fig:comp_discrete}) .
 %----------------------------------------
 \begin{table*}
 \centering
\caption{Summary of the proposed methods and other EAs assessed to the {\textbf{NYTP}},*CMA-ES results are feasible in terms of pipe sizes and nodal head pressure.}
\scalebox{0.7}{
\begin{tabular}{p{3.3cm}?p{1.2cm}|p{1.6cm}|p{2.3cm}|p{1.6cm}|p{2.4cm}|p{2cm}}
\hlineB{4}
\textbf{Algorithm} 
&\textbf{Number of runs} 
& \textbf{Best solution  (\$ M)}
& \textbf{Success rate (\%) (Global Optimum) }
& \textbf{Average Cost (\$ M)}
& \textbf{Average evaluations to discover the first best solution}
&\textbf{Maximum number of evaluations}
\\ \hlineB{4} \hline
  \textbf{SDE} \cite{zheng2012performance}
    &100   
  &38.64
  &97\%
  &38.65
  &$1.29\times 10^4$
  &$2.0\times 10^5$
       \\  \hline
   \textbf{DDE} \cite{zheng2012performance}
    &100   
  &38.64
  &93\%
  &38.66
  &$1.32\times 10^4$
  &$2.0\times 10^5$
       \\  \hline
  \textbf{SADE} \cite{zheng2012self}
  &50
  &38.64
  &92\%
  &38.64
  &$0.66\times 10^4$ 
  &NA
       \\  \hline
  
  \textbf{GHEST} \cite{bolognesi2010genetic}
  &60 
  &38.64 
  &92\%
  &38.64
  &$1.15\times 10^4$
  &NA
          \\ \hline
  
 \textbf{HD-DDS} \cite{tolson2009hybrid}
  &50 
  &38.64
  &86\%
  &38.64
  &$4.70\times 10^4$
  &$0.5\times 10^5$
       \\  \hline
   \textbf{DE} \cite{suribabu2010differential}
   &300 
  &38.64
  &71\%
  &NA
  &$0.55\times 10^4$
  &$1.0\times 10^5$
      \\  \hline
   \textbf{Scatter Search} \cite{lin2007scatter}
  &100   
  &38.64
  &65\%
  &NA
  &$5.76\times 10^4$
  &NA
      \\  \hline
    \textbf{MMAS} \cite{zecchin2007ant}
  &20 
  &38.64
  &60\%
  &38.64
  &$3.07\times 10^4$
  &$0.5\times 10^5$
      \\  \hline
   \textbf{CGA} \cite{zheng2012performance}
    &100   
  &38.64
  &50\%
  &39.04
  &$4.43\times 10^4$
  &$2.0\times 10^5$
       \\  \hline
        \textbf{SGA} \cite{zheng2012performance}
    &100   
  &38.64
  &45\%
  &39.25
  &$5.48\times 10^4$
  &$2.0\times 10^5$
       \\  \hline
        \textbf{PSO} \cite{montalvo2008particle}
    &  2000 
  &38.64
  &30\%
  &NA
  &NA
  &$1.0\times 10^5$
       \\  \hline
   \textbf{CMA-ES* }
  &
  & 
  &
  &
  &
  &
   \\
  $\lambda=10$&
  & 38.64 
  & 6.7\%
  &40.56
  &$0.63\times 10^4$
  &$1.0\times 10^5$
   \\
   $\lambda=20$&
  &38.64
  &3.3\%
  &41.97
  &$0.54\times 10^4$
  &$1.0\times 10^5$
   \\
   $\lambda=50$&30
  & 38.64
  & 3.3\%
  & 45.79
  &$0.72\times 10^4$
  &$1.0\times 10^5$
  \\
   $\lambda=100$&
  & 38.64
  & 6.7\%
  & 43.80
  & $1.01\times 10^4$
  &$1.0\times 10^5$
   \\
  $\lambda=200$&
  &38.64
  &13.3\%
  &42.92
  &$1.8\times 10^4$
  &$2.0\times 10^5$
   \\
  $\lambda=400$&
  &38.64
  &27.0\%
  &40.53
  &$2.1\times 10^4$
  &$2.0\times 10^5$
   \\
  \hline
   \textbf{CM$A_{ES}$-G$S_U$ }
  &
  & 
  &
  &
  & 
  &
  \\ 
   $\lambda=10$&
  &38.64
  &33.3\%
  &39.44
  &$0.64\times 10^4$
  &$1.0\times 10^5$
   \\
   $\lambda=20$&
  &38.64 
  &43.3\%
  &39.15
  &$0.55\times 10^4$
  &$1.0\times 10^5$
 \\
   $\lambda=50$
  &30
  &38.64
  &40.0\%
  &39.80
  &$0.73\times 10^4$
  &$1.0\times 10^5$
   \\
   $\lambda=100$&
  &38.64 
  &53.3\%
  &39.27
  &$1.1\times 10^4$
  &$1.0\times 10^5$
   \\
  $\lambda=200$&
  &38.64
  &83.3\%
  &38.85
  &$1.9\times 10^4$
  &$2.0\times 10^5$
   \\
  $\lambda=400$&
  &38.64
  &\textbf{100\%}
  &38.64
  &$2.2\times 10^4$
  &$2.0\times 10^5$
   \\
  \hline
   \textbf{CM$A_{ES}$-G$S_U$-G$S_D$ }
  &
  & 
  &
  &
  & 
  &
  \\ 
   $\lambda=10$&
  &38.64
  &36.6\%
  &39.30
  &$0.65\times 10^4$
  &$1.0\times 10^5$
   \\
   $\lambda=20$&
  &38.64 
  &43.3\%
  &39.10
  &$0.61\times 10^4$
  &$1.0\times 10^5$
 \\
   $\lambda=50$&30
  &38.64
  &43.3\%
  &39.49
  &$0.74\times 10^4$
  &$1.0\times 10^5$
   \\
   $\lambda=100$&
  &38.64 
  &56.7\%
  &39.19
  &$1.2\times 10^4$
  &$1.0\times 10^5$
   \\
  $\lambda=200$&
  &38.64
  &86.7\%
  &38.80
  &$2.0  \times 10^4$
  &$2.0\times 10^5$
   \\
   $\lambda=400$&
  &38.64
  &\textbf{100\%}
  &38.64
  &$2.3\times 10^4$
  &$2.0\times 10^5$
  \\
   \hline
    
   \textbf{RLS} 
  &
  & 
  &
  &
  &
  &
  \\
   $\sigma=0.1$&
  &53.42 
  &0.0\%
  &61.32
  &$1.44\times 10^4$ %14478
  &$1.0\times 10^5$
   \\
   $\sigma=0.25$&30
  &39.93 
  &0.0\%
  &56.26
  &$7.58\times 10^4$%75844
  &$1.0\times 10^5$
   \\
   $\sigma=0.5$&
  &39.52
  &0.0\%
  &55.99
  &$1.50\times 10^4$%15011
  &$1.0\times 10^5$
   \\
   $\sigma=Linear$&
  & 39.43
  &0.0\%
  &57.49
  &$8.50\times 10^4$%85069
  &$1.0\times 10^5$
   \\\hline
   \textbf{1+1EA}
  &
  & 
  &
  &
  &
  &
  \\
   $\sigma=0.1$&
  &44.04
  &0.0\%
  &50.93
  &$9.10\times 10^4$%91094
  &$1.0\times 10^5$
   \\
   $\sigma=0.25$&30
  &39.74 
  &0.0\%
  &49.76
  &$6.89\times 10^4$%68930
  &$1.0\times 10^5$
  \\
   $\sigma=0.5$&
  &38.88 
  &0.0\%
  &49.76
  &$5.86\times 10^4$%58620
  &$1.0\times 10^5$
  \\
   $\sigma=Linear$&
  &39.47
  &0.0\%
  &47.31
  &$5.96\times 10^4$%59640
  &$1.0\times 10^5$
   \\
  \hlineB{4}
\end{tabular}
}
\label{summary_per_NYTP}
\end{table*}
 %--------------------------
 %----------------------------------
 %------------------------------------------
 \subsection{Case Study 2: Doubled New York Tunnel problem (NYTP2)}
 
The NYTP2 consists of two independent NYTP hydraulically which are connected by one reservoir. The number of decision variables is 42, and the design options number is similar to NYTP. The best-known design cost is \$77.276 million. The best-founded results for NYTP2 were reported by a Self-Adaptive DE (SADE \cite{zheng2012self}). The SADE success rate for finding the best-known solution is \%90 of the time, and the average number of evaluations is 33810.

In this paper, a considerable performance of CM$A_{ES}$-G$S_U$ for NYTP2 is also observed, as CM$A_{ES}$-G$S_U$ is able to find the optimal solution in 100\% of the 30 independent runs with a uniform random scenario for initializing the decision variables (Table \ref{summary_per_DNYTP}). Moreover, CM$A_{ES}$-G$S_U$ can beat the CMA-ES easily in terms of both the quality and efficiency of the obtained solutions. Indeed, the CMA-ES performance can be proper if the right population size is chosen. For both case studies (NYTP and NYTP2), big population sizes ($\lambda=10,20\times Dimension$) have better performance because of having more robust exploration strength compared with the exploitation ability.

In spite of the simplicity of RLS and (1+1)EA, two methods can find the relative near-optimal designs at \$77.69 ($\sigma=0.5$) and \$79.38 million ($\sigma=Linear$) respectively. These corresponding closeness in solutions show that the RLS and (1+1)EA are exploring in the neighbourhood space of the known-optimum solution.
%--Summery of DNYTP------------------
 \begin{table*}
 \centering
\caption{Summary of the proposed methods and other EAs evaluated to the \textbf{DNYTP((NYTP2)}}
\scalebox{0.7}{
\begin{tabular}{p{3.3cm}?p{1.2cm}|p{1.6cm}|p{2.3cm}|p{1.6cm}|p{2.4cm}|p{1.6cm}}
\hlineB{4}
\textbf{Algorithm} 
&\textbf{Number of runs} 
& \textbf{Best solution  (\$ M)}
& \textbf{Success rate (\%) (Founded Best Solution) }
& \textbf{Average Cost (\$ M)}
& \textbf{Average evaluations to discover the first best solution}
&\textbf{Maximum number of evaluations}
\\ \hlineB{4} \hline
  \textbf{SADE} \cite{zheng2012self}
  &50
  &77.28
  &90\%
  &77.28
  &$0.34\times 10^5$ 
  &NA
       \\  \hline

 \textbf{HD-DDS} \cite{tolson2009hybrid}
  &20 
  &77.28
  &85\%
  &77.28
  &$3.10\times 10^5$
  &$3.1\times 10^5$
       \\  \hline
   \textbf{DE} \cite{zheng2012self}
  &50 
  &77.28
  &86\%
  &77.28
  &$0.70\times 10^5$
  &$1.0\times 10^5$
      \\  \hline
   
    \textbf{MMAS} \cite{zecchin2007ant}
  &20 
  &77.28
  &5\%
  &78.20
  &$2.38\times 10^5$
  &$3.0\times 10^5$
      \\  \hline
      \textbf{CMA-ES }
  &
  & 
  &
  &
  &
  &
   \\
  $\lambda=10$&
  &78.04
  &0.0\%
  &84.17
  &$0.18\times 10^5$
  &$1.0\times 10^5$
   \\
   $\lambda=20$&
  & 81.15
  &0.0\%
  &85.11
  &$0.19\times 10^5$
  &$1.0\times 10^5$
   \\
   $\lambda=50$&30
  &81.73
  &0.0\%
  &89.23
  &$0.23\times 10^5$
  &$1.0\times 10^5$
  \\
   $\lambda=100$&
  & 77.28 
  &3.33\%
  & 93.05
  &$0.32\times 10^5$
  &$1.0\times 10^5$
   \\
  $\lambda=200$&
  &77.28
  &3.33\%
  &96.45
  &$0.48\times 10^5$
  &$2.0\times 10^5$
   \\
  $\lambda=400$&
  &77.28
  &3.33\%
  &93.44
  &$0.71\times 10^5$
  &$2.0\times 10^5$
   \\
    $\lambda=600$&
  &77.28
  &3.33\%
  &86.11
  &$0.92\times 10^5$
  &$2.0\times 10^5$
   \\
    $\lambda=800$&
  &79.74
  &0.0\%
  &85.87
  &$0.98\times 10^5$
  &$2.0\times 10^5$
   \\
    $\lambda=1000$&
  &77.28
  &6.66\%
  &80.20
  &$1.01\times 10^5$
  &$2.0\times 10^5$ \\
  \hline
   \textbf{CM$A_{ES}$-G$S_U$ }
  &
  & 
  &
  &
  & 
  &
  \\ 
   $\lambda=10$&
  &77.45
  &0.0\%
  &79.76
  &$0.19\times 10^5$
  &$1.0\times 10^5$
   \\
   $\lambda=20$&
  &77.28
  &13.33\%
  &80.05
  &$0.20\times 10^5$
  &$1.0\times 10^5$
 \\
   $\lambda=50$
   &30
  &77.28
  &13.33\%
  &79.53
  &$0.24\times 10^5$
  &$1.0\times 10^5$
   \\
   $\lambda=100$&
  &77.28 
  &13.33\%
  &79.50
  &$0.33\times 10^5$
  &$1.0\times 10^5$
   \\
  $\lambda=200$&
  &77.28
  &23.33\%
  &79.48
  &$0.49\times 10^5$
  &$2.0\times 10^5$
   \\
  $\lambda=400$&
  &77.28
  &33.33\%
  &78.93
  &$0.72\times 10^5$
  &$2.0\times 10^5$
   \\
    $\lambda=600$&
  &77.28
  &70.00\%
  &77.82
  &$0.93\times 10^5$
  &$2.0\times 10^5$
   \\
    $\lambda=800$&
  &77.28
  &90.00\%
  &77.45
  &$0.99\times 10^5$
  &$2.0\times 10^5$
   \\
   $\lambda=1000$&
  &77.28
  &\textbf{100\%}
  &77.28
  &$1.03\times 10^5$
  &$2.0\times 10^5$\\
  \hline
   \textbf{CM$A_{ES}$-G$S_U$-G$S_D$ }
  &
  & 
  &
  &
  & 
  &
  \\ 
   $\lambda=10$&
  &77.45
  &0.0\%
  &79.56
  &$0.20\times 10^5$
  &$1.0\times 10^5$
   \\
   $\lambda=20$&
  &77.28
  &16.67\%
  &79.77
  &$0.21\times 10^5$
  &$1.0\times 10^5$
 \\
   $\lambda=50$&30
  &77.28
  &13.33\%
  &79.18
  &$0.25\times 10^5$
  &$1.0\times 10^5$
   \\
   $\lambda=100$&
  &77.28 
  &13.33\%
  &79.28
  &$0.34\times 10^5$
  &$1.0\times 10^5$
   \\
  $\lambda=200$&
  &77.28
  &23.33\%
  &79.20
  &$0.50\times 10^5$
  &$2.0\times 10^5$
   \\
  $\lambda=400$&
  &77.28
  &36.66\textbf{\%}
  &78.56
  &$0.73\times 10^5$
  &$2.0\times 10^5$
   \\
    $\lambda=600$&
  &77.28
  &73.33\%
  &77.71
  &$0.94\times 10^5$
  &$2.0\times 10^5$
   \\
    $\lambda=800$&
  &77.28
  &90.00\%
  &77.44
  &$1.01\times 10^5$
  &$2.0\times 10^5$
   \\
   $\lambda=1000$&
  &77.28
  &\textbf{100\%}
  &77.28
  &$1.04\times 10^5$
  &$2.0\times 10^5$\\
  \hline
   \textbf{RLS} 
  &
  & 
  &
  &
  &
  &
  \\
   $\sigma=0.1$&
  &106.92
  &0.0\%
  &122.64
  &$0.26\times 10^5$%26060
  &$1.5\times 10^5$
   \\
   $\sigma=0.25$&30
  &79.87
  &0.0\%
  &112.53
  &$0.13\times 10^5$%12893
  &$1.5\times 10^5$
   \\
   $\sigma=0.5$&
  &77.69
  &0.0\%
  &111.99
  &$0.29\times 10^5$%28520
  &$1.5\times 10^5$
  \\
   $\sigma=Linear$&
  &78.87
  &0.0\%
  &114.98
  &$1.5\times 10^5$
  &$1.5\times 10^5$
   \\\hline
   \textbf{1+1EA}
  &
  & 
  &
  &
  &
  &
  \\
   $\sigma=0.1$&
  &88.09
  &0.0\%
  &101.88
  &$1.46\times 10^5$%145750
  &$1.5\times 10^5$
   \\
   $\sigma=0.25$&30
  & 79.48
  &0.0\%
  &99.53
  &$1.17\times 10^5$%117181
  &$1.5\times 10^5$
  \\
   $\sigma=0.5$&
  & 79.49
  &0.0\%
  &98.40
  &$1.11\times 10^5$%111378
  &$1.5\times 10^5$
  \\
   $\sigma=Linear$&
  &79.38
  &0.0\%
  &94.63
  &$1.01\times 10^5$%101388
  &$1.5\times 10^5$
   \\
  \hlineB{4}
\end{tabular}
}
\label{summary_per_DNYTP}
\end{table*}
%------------------------------------
\subsection{Case Study 3: 50 * New York Tunnel problem (NYTP50)}
The NYTP50 \cite{zheng2011optimal} includes  50 individualistic NYTP in terms of hydraulical equations that are joined by one tank. This problem can be a large-scale optimization benchmark that decision variables number is $50 \times 21$ with the same design options of the original NYTP. The best-known design cost can be calculated at \$1932 million. 

Table \ref{summary_per_50NYTP} shows the best and average founded design costs are \$2022(M) and \$2030(M) by CM$A_{ES}$-G$S_U$-G$S_D$ of the ten independent runs with a uniform random scenario for initializing the decision variables where the population size is 500. These relative near-optimal designs are considerably better than the 50NYTP designs from \cite{zheng2011optimal} that represents the proposed hybrid method is able to explore in a vast search space properly by a few iterations compared with the decision variable length and previous optimization methods. 
%--Summery of DNYTP------------------
 \begin{table*}
 \centering
\caption{Summary of the proposed methods and other EAs evaluated to the $50\times NYTP$ (NYTP50), the global best=$\$1932(M)$.( * shows the designs are continuous  and the average pressure violation per each NYTP is 0.09, 0.15 and 0.2 where $\lambda=200,500$ and $1000$)respectively.}
\scalebox{0.8}{
\begin{tabular}{p{3.4cm}?p{1.2cm}|p{1.6cm}|p{2.3cm}|p{1.6cm}|p{2.4cm}|p{1.8cm}}
\hlineB{4}
\textbf{Algorithm} 
&\textbf{Number of runs} 
& \textbf{Best solution  (\$ M)}
& \textbf{Success rate (\%) (Founded Best Solution) }
& \textbf{Average Cost (\$ M)}
& \textbf{Average evaluations to discover the first best solution}
&\textbf{Maximum number of evaluations}
\\ \hlineB{4} \hline
  \textbf{GA} \cite{zheng2011optimal}
  &100
  &2238
  &0.0\%
  &2321
  & NA
  &$40.0\times10^6$
       \\  \hline
      \textbf{CMA-ES }
  &
  & 
  &
  &
  &
  &
   \\
  $\lambda=200$
  &10
  &1873*
  &0.0\%
  &1897*
  &$0.258\times 10^6$%258000
  &$1.0\times 10^6$
   \\
   $\lambda=500$
   &10
  &1824*
  &0.0\%
  &1834*
  &$0.636\times 10^6$%636000
  &$1.0\times 10^6$
   \\
   $\lambda=1000$
   &10
  &1814*
  &0.0\%
  &1817*
  &$0.912\times 10^6$%912000
  &$1.0\times 10^6$
  \\ \hline
   
   \textbf{CM$A_{ES}$-G$S_U$ }
  &
  & 
  &
  &
  & 
  &
  \\ 
   $\lambda=200$
   &10
  &2033
  &0.0\%
  &2037
  &$0.436\times 10^6$%463000
  &$1.0\times 10^6$
   \\
   $\lambda=500$
   &10
  &2026
  &0.0\%
  &2037
  &$0.752\times 10^6$%752000
  &$1.0\times 10^6$
 \\
   $\lambda=1000$
   &10
  &2068
  &0.0\%
  &2090
  &$0.942\times 10^6$%942000
  &$1.0\times 10^6$
   \\
     \hline
   \textbf{CM$A_{ES}$-G$S_U$-G$S_D$ }
  &
  & 
  &
  &
  & 
  &
  \\ 
   $\lambda=200$
   &10
  &2030
  &0.0\%
  &2033
  &$0.485\times 10^6$%485000
  &$1.0\times 10^6$
   \\
   $\lambda=500$
   &10
  &2022
  &0.0\%
  &2030
  &$0.771\times 10^6$%771000
  &$1.0\times 10^6$
 \\
   $\lambda=1000$
   &10
  &2055
  &0.0\%
  &2069
  &$0.981\times 10^6$%981000
  &$1.0\times 10^6$
  \\
  \hlineB{4}
\end{tabular}
}
\label{summary_per_50NYTP}
\end{table*}
%------------------------------------
%--------------------------------------
\subsection{Case Study 4: Hanoi (HP) }
The Hanoi Network (HP) is made up of 34 pipes, 32 nodes, and three loops.  A gravity-fed system has been designed which is fed from a single fixed tank and is produced to fulfil assigned demands at the necessary pressures. The decision number of the problem can be six sizes of the industrial pipe diameters are possible, and also the cost of $i^{th}$ pipe with diameter $D_i$ and particular length $L_i$ can be computed by the formula ($C_{pipe_i}= 1.1 \times D^{1.5}_i\times L_i$). Where the pipe diameter is in inches, and the pipe length is in meters. The Hazen-Williams coefficient is deterministic at 130 for total pipes. All required data can be obtained from the reference \cite{fujiwara1990two}. The best-founded feasible solution for the optimization of the Hanoi network cost is  \$6.081 million; it is referred to the literature. The Hanoi Problem has been taken into account as three different aspects of optimization problem such as a continuous \cite{fujiwara1990two}, split-pipe \cite{fujiwara1990two}, and discrete pipes  \cite{savic1997genetic}, \cite{wu2001using}, \cite{cunha1999water}.  Some of the best achieved HP layouts, which are introduced by the authors are listed in Table \ref{table_review_HP}. It can be seen; where the pipe diameters are considered as continuous, the best performance is allocated to the CMA-ES which is able to find the cheapest feasible continuous solution by \$5.959 million. On the other hand,  most of the researchers have been focused on the discrete pipe sizes recently, so in Table \ref{summary_per_Hanoi}, the discrete results are reported only.

%--------------------------------------
%  \begin{figure}
%  \centering
% \includegraphics[ width=0.7\textwidth]
% {HP_con.eps}
%  \caption{ The comparison of the average convergence rate of CMA-ES ,RLS and 1+1EA for the HP design where  $\lambda=10, 20, 50, 100, 200, 500$ and $\sigma=0.1, 0.25, 0.5, Linear.$ } \label{fig:HP_convergence}
%  \end{figure}
 %------------------------------------
 \begin{figure*}[t]%[H]
\centering

\subfloat[]{
\includegraphics[clip,width=0.6\columnwidth]{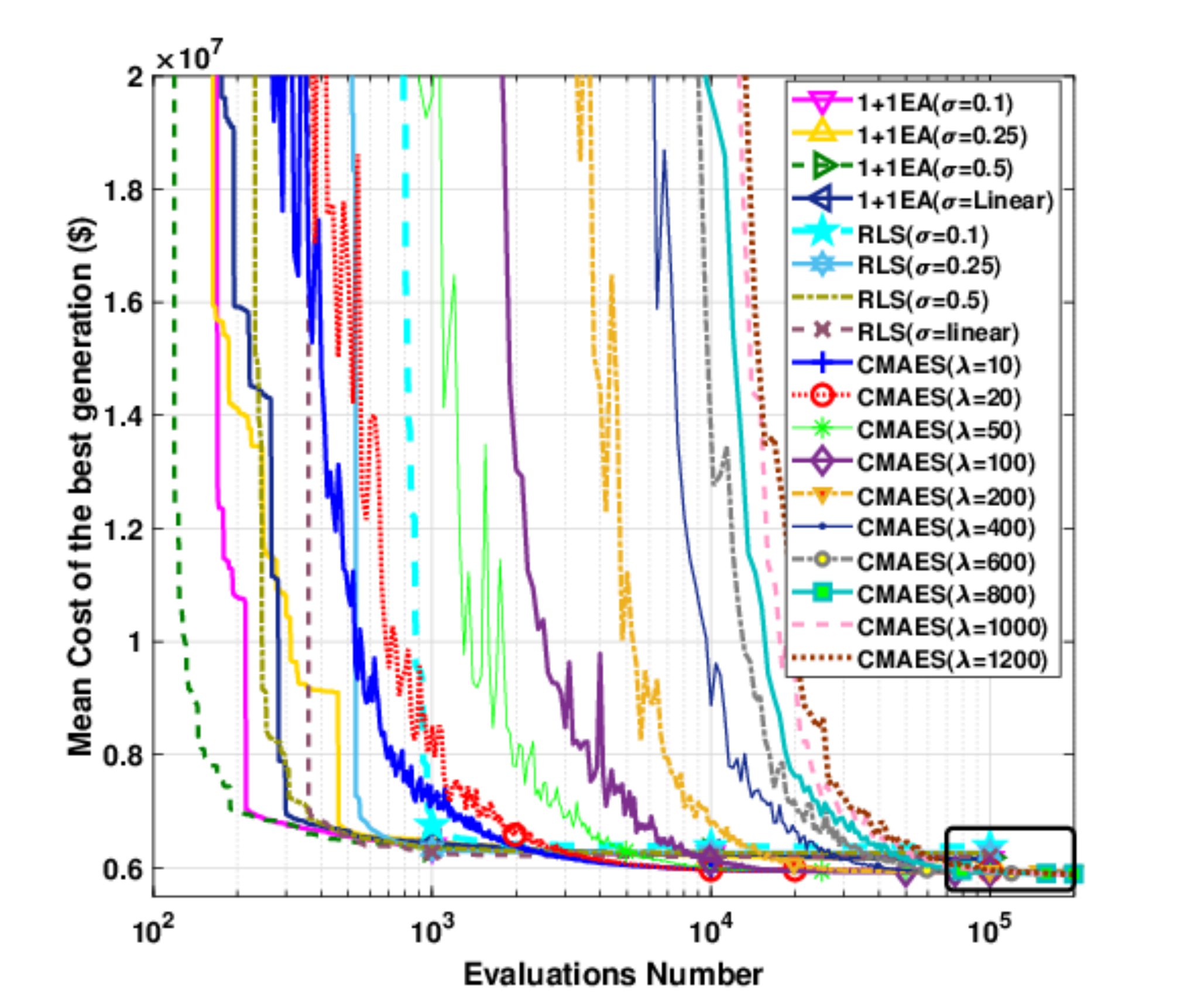}}
\subfloat[]{
\includegraphics[clip,width=0.4\columnwidth]{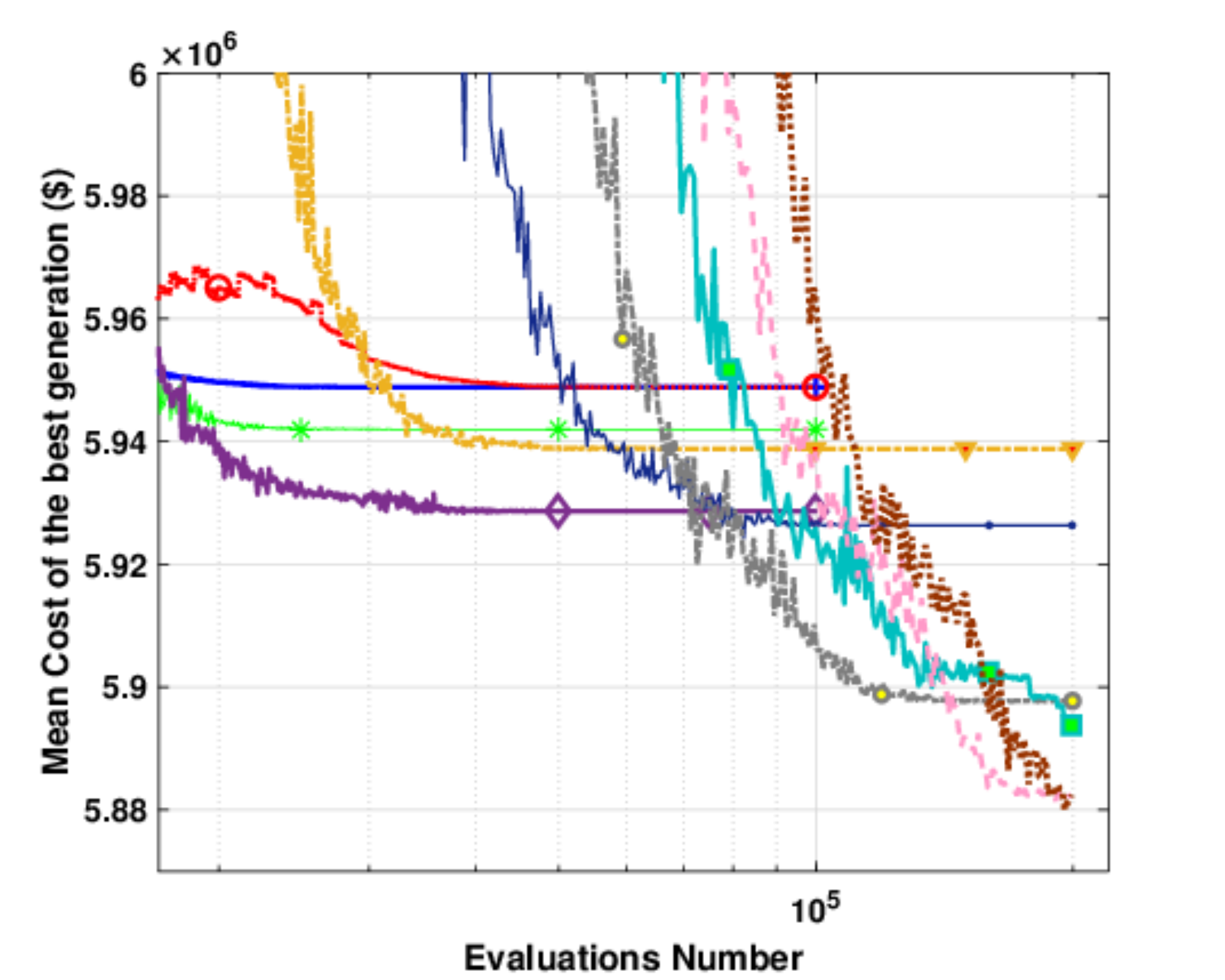}}\\

\caption{The comparison of the average convergence rate of CMA-ES ,RLS and 1+1EA for the HP design where  $\lambda=10, 20, 50, 100, 200, 400, 600, 800, 1000, 1200$ and $\sigma=0.1, 0.25, 0.5, Linear.$ (a) and a zoomed version of the last iterations is shown (b)}%
\label{fig:HP_con}%
\end{figure*}
%-------------------------------
According to the previous optimization results of case studies (NYTP, NYTP2 and 50NYTP), the efficiency of the CMA-ES with the continuous decision variables is better than discrete. A comparison of CMA-ES effectiveness with discrete pipe sizes can be illustrated in Figure \ref{fig:Dis_Hanoi} by diverse population sizes for minimizing the pipe cost of HP. The large population size found better solutions similar to previous case studies, but as a matter of fact, the continuous version of CMA-ES is more robust to explore the search space and find the optimal solutions which are placed very near the constraint edges. Thus initially, CMA-ES is applied by the continuous pipe sizes and then obtained solutions are rounded to commercialized pipe sizes. As a result, a set of near-optimal feasible layouts is exploited by the CMA-ES (such as \$6.173, \$6.204 million) and the convergence rate is considerable. However, the best-known discrete solution is not found by the CMA-ES. Promisingly, CMA-ES can discover a bunch of very cheap continuous layouts of the HP that some of them are feasible or infeasible in terms of nodal pressure head violations. This is the primary motivation for applying the next step. For improving the optimization process, two other parts of the proposed hybrid method are employed. Initially, $GS_U$ tries to fix up the violation of the nodal pressure head by increasing the size of pipes discretely. Secondly, both ideas $GS_D$ is evaluated independently to reduce the extra imposed pipe cost of the  $GS_U$ step. Interestingly, the proposed hybrid method is able to find the well-known HP design 83.33\% of the time over 30 independent runs with different initializing random number seeds, when the population size is 1000.

 Figure \ref{fig:NYTP_convergence} describes the average convergence rate of the RLS, 1+1EA and CMAES methods by 30 independent runs. The 1+1EA convergence speed is faster than others with $\sigma=0.5$ and 0.1. Moreover,  CMAES with small population sizes can converge quicker than big population sizes. However, the big population sizes performance for exploiting the search space are more considerable finally.

Meanwhile, using the third step of the hybrid framework ($GS_D$) leads to increasing the percentage of finding the optimum solution (success rate) to 19\% compared with the CM$A_{ES}$-G$S_U$with spending just 8.2\% more computational budgets.

 According to the observed results from the Table \ref{summary_per_Hanoi}, RLS and (1+1)EA are able to find semi-optimum solutions means very close to the best-known solution. The performance of both RLS and (1+1)EA are acceptable based on the expectations; however, the best-introduced solution of HP is not found by them. The (1+1)EA efficiency (\$6.115 million, ($\sigma=0.25$) ) can be better than RLS (\$6.128 million, ($\sigma=0.1$)) in terms of the quality achieved solution.  However, the average performance of RLS with regard to the mean evaluation number and quality to find the best solution can overcome the (1+1)EA.  Where the successful mutation step sizes are small at $0.1$ or $0.25$, are declared the search space is a highly complex multi-modal, so a robust exploitation technique needs for the HP case study.  Interestingly, the (1+1)EA, which is a simple EAs can beat a complex method like MMAS \cite{zecchin2007ant}. 
 
 \begin{figure}
 \centering
\includegraphics[ width=0.7\textwidth]
{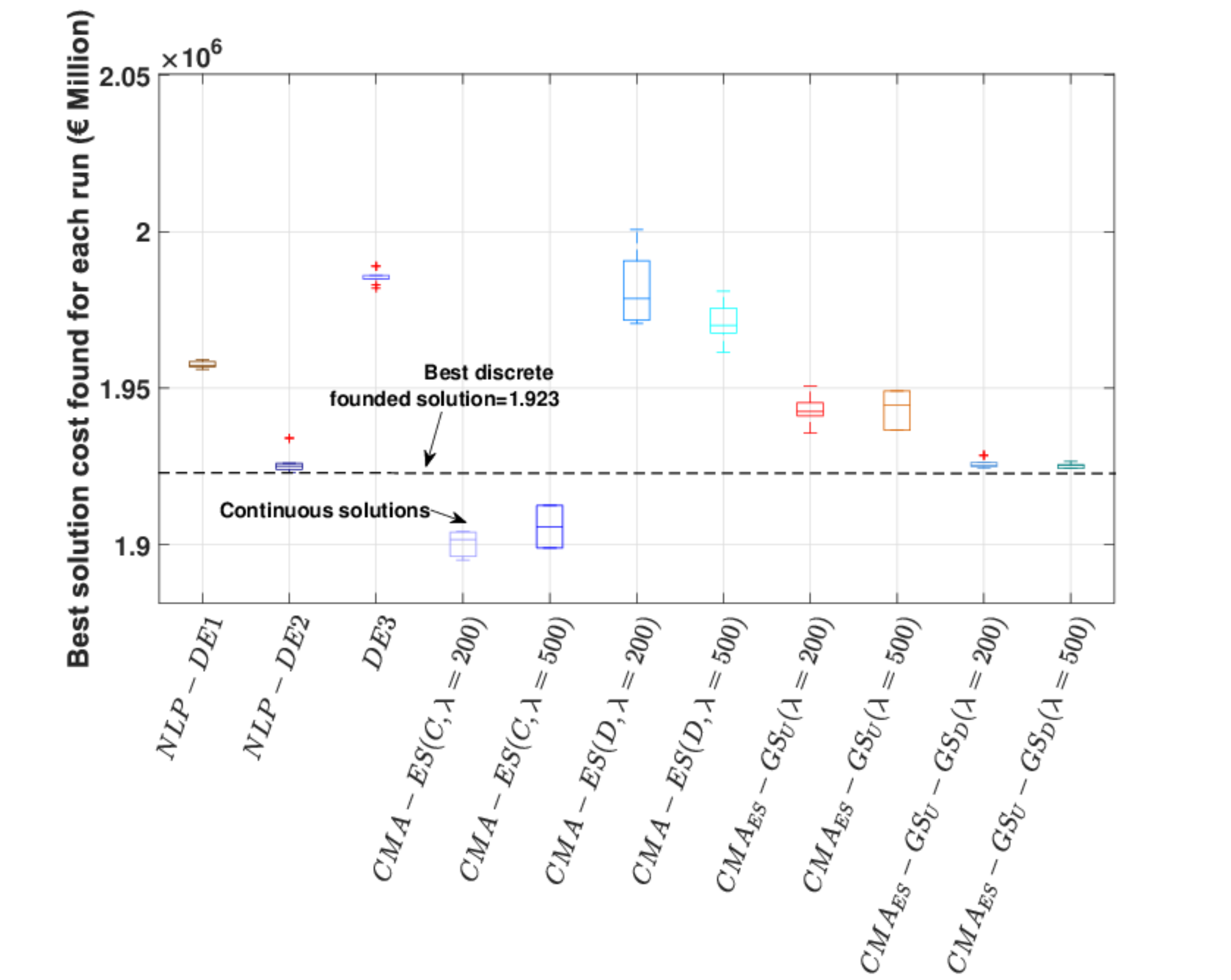}
 \caption{ The efficiency comparison of proposed Hybrid framework: CMA-ES: Continuous and Discrete($\lambda=200,500$) , CM$A_{ES}$-G$S_U$ and CM$A_{ES}$-G$S_U$-G$S_D$, with previous best methods \cite{zheng2011combined} ($NLP-DE_1$, $NLP-DE_2$ and $DE_3$) for {Balerma Network}. The average performance of CM$A_{ES}$-G$S_U$-G$S_D$ is better than the previous best method ($NLP-DE_2$).  } \label{fig:BN_boxplot_quality}
 \end{figure}
 %------------------------------------
 
%--summery of Hanoi------------------
 \begin{table*}
 \centering
\caption{Summary of the proposed methods and other EAs evaluated to the {\textbf{Hanoi}}}
\scalebox{0.8}{
\begin{tabular}{p{3.3cm}?p{1.2cm}|p{1.6cm}|p{2.0cm}|p{1.6cm}|p{2.4cm}|p{1.4cm}}
\hlineB{4}
\textbf{Algorithm} 
&\textbf{Number of runs} 
& \textbf{Best solution  (\$ M)}
& \textbf{Success rate (\%) }
& \textbf{Average Cost (\$ M)}
& \textbf{Average evaluations to discover the first best solution}
&\textbf{Maximum number of evaluations}
\\ \hlineB{4} \hline
  \textbf{BLP-DE} \cite{zheng2013coupled}
  &100
  &6.081
  &98\%
  &6.085
  &$3.31\times 10^4$ 
  &$0.4\times 10^5$
       \\  \hline
        \textbf{NLP-DE} \cite{zheng2013coupled}
  &100
  &6.081
  &97\%
  &6.082
  &$3.46\times 10^4$ 
  &$0.8\times 10^5$
       \\  \hline
        \textbf{SDE} \cite{zheng2012performance}
    &100   
  &6.081
  &92\%
  &NA
  &$7.72\times 10^4$
  &$5.0\times 10^5$
       \\  \hline
  \textbf{SADE} \cite{zheng2012self}
  &50
  &6.081
  &84\%
  &6.090
  &$6.05\times 10^4$ 
  &$2.0\times 10^5$
       \\  \hline
    \textbf{DE} \cite{suribabu2010differential}
  &300 
  &6.081
  &80\%
  &NA
  &$4.87\times 10^4$
  &$1.0\times 10^5$
      \\  \hline
      \textbf{DDE} \cite{zheng2012performance}
    &100   
  &6.081
  &80\%
  &NA
  &$6.37\times 10^4$
  &$5.0\times 10^5$
       \\  \hline
       \textbf{Scatter Search} \cite{lin2007scatter}
  &100   
  &6.081
  &64\%
  &NA
  &$4.31\times 10^4$
  &NA
      \\  \hline
       \textbf{GHEST} \cite{bolognesi2010genetic}
  &60 
  &6.081 
  &38\%
  &6.175
  &$5.01\times 10^4$
  &NA
          \\ \hline
  \textbf{GENOME} \cite{reca2006genetic}
   &10 
  &6.081
  &10\%
  &6.248
  &NA
  &$1.5\times 10^5$
      \\  \hline
 \textbf{HD-DDS} \cite{tolson2009hybrid}
  &50 
  &6.081
  &8\%
  &6.252
  &$10.00\times 10^4$
  &$1.0\times 10^5$
       \\  \hline
           
        \textbf{PSO} \cite{montalvo2008particle}
  &2000 
  &6.081
  &5\%
  &6.310
  &NA
  &$0.8\times 10^5$
       \\  \hline
        \textbf{CGA} \cite{zheng2012performance}
    &100   
  &6.109
  &0.0\%
  &6.274
  &$32.12\times 10^4$
  &$5.0\times 10^5$
       \\  \hline
        \textbf{SGA} \cite{zheng2012performance}
    &100   
  &6.112
  &0.0\%
  &6.287
  &$3.85\times 10^4$
  &$5.0\times 10^5$
       \\  \hline
        \textbf{MMAS} \cite{zecchin2007ant}
  &20 
  &6.134
  &0.0\%
  &6.386
  &$8.50\times 10^4$
  &$1.0\times 10^5$
      \\  \hlineB{2}
      \textbf{CMA-ES }
  &
  & 
  &
  &
  &
  &
   \\
  $\lambda=10$&
  &7.371
  &0.0\%
  &7.573
  &$0.73\times 10^4$
  &$1.0\times 10^5$
   \\
   $\lambda=20$&
  &6.840
  &0.0\%
  &6.950
  &$0.72\times 10^4$
  &$1.0\times 10^5$
   \\
   $\lambda=50$&
  &6.733
  &0.0\%
  &6.931 
  &$0.96\times 10^4$
  &$1.0\times 10^5$
  \\
   $\lambda=100$&
  &6.494 
  &0.0\%
  &6.695 
  &$1.41\times 10^4$ 
  &$1.0\times 10^5$
   \\
  $\lambda=200$&30
  &6.281
  &0.0\%
  &6.295
  &$2.28\times 10^4$ 
  &$2.0\times 10^5$
   \\
  $\lambda=400$&
  &6.220
  &0.0\%
  &6.319
  &$4.70\times 10^4$ 
  &$2.0\times 10^5$
   \\
    $\lambda=600$&
  &6.227
  &0.0\%
  &6.287
  &$8.05\times 10^4$ 
  &$2.0\times 10^5$
  \\
   $\lambda=800$&
  &6.290
  &0.0\%
  &6.315
  &$1.01\times 10^5$
  &$2.0\times 10^5$
  \\
   $\lambda=1000$&
  &6.210
  &0.0\%
  &6.279
  &$1.39\times 10^5$
  &$2.0\times 10^5$
  \\
   $\lambda=1200$&
  &6.129
  &0.0\%
  &6.294
  &$1.6\times 10^5$
  &$2.0\times 10^5$
  \\
  \hlineB{2}

{CM$A_{ES}$-G$S_U$ }
  &
  & 
  & 
  &
  & 
  &
  \\
   $\lambda=10$&
  &6.094
  &0.0\%
  &6.293
  &$0.80\times 10^4$
  &$1.0\times 10^5$
   \\
   $\lambda=20$&
  &6.224
  &0.0\%
  &6.296
  &$0.79\times 10^4$
  &$1.0\times 10^5$
 \\
   $\lambda=50$
   &
  &6.199
  &0.0\%
  &6.276
  &$1.05\times 10^4$
  &$1.0\times 10^5$
   \\
   $\lambda=100$&
  &6.093
  &0.0\%
  &6.250
  &$1.48\times 10^4$
  &$1.0\times 10^5$
   \\
  $\lambda=200$&30
  &6.196
  &0.0\%
  &6.247
  &$2.47\times 10^4$ 
  &$2.0\times 10^5$
   \\
  $\lambda=400$&
  &6.081
  &6.67\%
  &6.219
  &$4.81\times 10^4$ 
  &$2.0\times 10^5$
   \\
   $\lambda=600$&
  &6.081
  &10.00\%
  &6.156
  &$8.16\times 10^4$ 
  &$2.0\times 10^5$
  \\
  $\lambda=800$&
  &6.081
  &13.33\%
  &6.149
  &$1.05\times 10^5$
  &$2.0\times 10^5$
  \\
  $\lambda=1000$&
  &6.081
  &36.67\%
  &6.112
  &$1.42\times 10^5$
  &$2.0\times 10^5$
  \\
  $\lambda=1200$&
  &6.081
  &33.33\%
  &6.118
  &$1.71\times 10^5$
  &$2.0\times 10^5$
  \\
  \hlineB{2}
   {CM$A_{ES}$-G$S_U$-G$S_D$ }
  &
  & 
  & 
  &
  & 
  &
  \\ 
   $\lambda=10$&
  &6.081
  &6.67\%
  &6.272
  &$0.83\times 10^4$
  &$1.0\times 10^5$
   \\
   $\lambda=20$&
  &6.224
  &0.0\%
  &6.282
  &$0.81\times 10^4$
  &$1.0\times 10^5$
 \\
   $\lambda=50$&
  &6.192
  &0.0\%
  &6.265
  &$1.12\times 10^4$
  &$1.0\times 10^5$
   \\
   
   $\lambda=100$&
  &6.081
  &3.33\%
  &6.241
  &$1.53\times 10^4$
  &$1.0\times 10^5$
   \\
  $\lambda=200$&30
  &6.192
  &0.0\%
  &6.237
  &$2.50\times 10^4$ 
  &$2.0\times 10^5$
   \\
  $\lambda=400$&
  &6.081
  &16.67\%
  &6.207
  &$4.91\times 10^4$ 
  &$2.0\times 10^5$
   \\
    $\lambda=600$&
  &6.081
  &43.33\%
  &6.150
  &$8.28\times 10^4$ 
  &$2.0\times 10^5$
  \\
  $\lambda=800$&
  &6.081
  &53.33\%
  &6.142
  &$1.07\times 10^5$
  &$2.0\times 10^5$
  \\
  $\lambda=1000$&
  &6.081
  &83.33\%
  &6.106
  &$1.48\times 10^5$
  &$2.0\times 10^5$
  \\
  $\lambda=1200$&
  &6.081
  &80.00\%
  &6.110
  &$1.82\times 10^5$
  &$2.0\times 10^5$
  \\
  \hlineB{2}
   \textbf{RLS} 
  &
  & 
  &
  &
  &
  &
  \\
   $\sigma=0.1$&
  &6.128
  &0.0\%
  &8.355
  &$7.45\times 10^4$%74492
  &$1.0\times 10^5$
   \\
   $\sigma=0.25$&30
  & 6.170
  &0.0\%
  &7.123
  &$4.82\times 10^4$%48204
  &$1.0\times 10^5$
   \\
   $\sigma=0.5$&
  & 6.143
  &0.0\%
  &7.863
  &$6.71\times 10^4$%67120
  &$1.0\times 10^5$
  \\
   $\sigma=Linear$&
  & 6.136
  &0.0\%
  &8.286
  &$6.12\times 10^4$%61185
  &$1.0\times 10^5$
   \\\hlineB{2}
   \textbf{1+1EA}
  &
  & 
  &
  &
  &
  &
  \\
   $\sigma=0.1$&
  &6.134
  &0.0\%
  & 9.39
  &$8.55\times 10^4$%85478
  &$1.0\times 10^5$
   \\
   $\sigma=0.25$&30
  &6.115
  &0.0\%
  &9.184
  &$8.88\times 10^4$%88840
  &$1.0\times 10^5$
  \\
   $\sigma=0.5$&
  & 6.202
  &0.0\%
  & 9.167
  &$8.83\times 10^4$%88300
  &$1.0\times 10^5$
  \\
   $\sigma=Linear$&
  &6.170
  &0.0\%
  &9.045
  &$9.84\times 10^4$%98367
  &$1.0\times 10^5$
   \\
  \hlineB{4}
\end{tabular}
}
\label{summary_per_Hanoi}
\end{table*}
%-------------------------------------
%--------BN network-----------------
 \begin{table*}
 \centering
\caption{Summary of the proposed methods and other EAs evaluated to the \textbf{{(BN)}}}
\scalebox{0.7}{
\begin{tabular}{|p{3.7cm}|p{1.2cm}|p{1.5cm}|p{1.3cm}|p{1.4cm}|p{2.5cm}|p{1.7cm}|}
\hline
\textbf{Algorithm} 
&\textbf{Number of runs} 
& \textbf{Best ~~~ solution  (\euro M)}
& \textbf{Success rate(\%) }
& \textbf{Average Cost ~~(\euro M)}
& \textbf{Average evaluations to discover the first best~~ solutions}
&\textbf{Maximum number of evaluations}
\\ \hline \hline
\textbf{NLP-DE2} \cite{zheng2011combined}
  & 10
  &1.923
  &10.0\%
  &1.927
  &$1.428\times 10^6$
  &$2.0\times 10^6$
  \\  \hline
  \textbf{HD-DDS-1} \cite{tolson2009hybrid}
  &1 
  &1.941
  &0.0\%
  &NA
  &$30.00 \times 10^6$
  &$30.0\times 10^6$
  \\  \hline
\textbf{NLP-DE1} \cite{zheng2011combined}
  & 10
  &1.956
  &0.0\%
  &1.957
  &$4.12\times 10^3$
  &$1.0\times 10^6$
  \\  \hline
  \textbf{HD-DDS-2} \cite{tolson2009hybrid}
  &10 
  &1.956
  &0.0\%
  &NA
  &$30.00 \times 10^6$
  &$10.0\times 10^6$
  \\  \hline
    \textbf{DE3} \cite{zheng2011combined}
  & 10
  &1.982
  &0.0\%
  &1.986
  &$9.21\times 10^6$
  &$10.0\times 10^6$
  \\  \hline
  \textbf{SADE} \cite{zheng2012self}
  &10
  &1.983
  &0.0\%
  &1.995
  &$1.20\times10^6$ 
  &$1.3\times 10^6$
       \\  \hline
       \textbf{CSHS} \cite{sheikholeslami2016hybrid}
  &10
  &1.988
  &0.0\%
  &2.031
  &$3.00\times10^6$ 
  &$5.0\times 10^6$
       \\  \hline
   \textbf{DE} \cite{zheng2012self}
  &10 
  &1.998
  &0.0\%
  &2.031
  &$2.30\times10^6$ 
  &$2.4\times 10^6$
      \\  \hline
   
    \textbf{GHEST} \cite{bolognesi2010genetic}
   &10 
  &2.002
  &0.0\%
  &2.055
  &$0.25\times10^6$ 
  &$10.0\times 10^6$
      \\  \hline
       \textbf{HS} \cite{geem2009particle}
   &NA 
  &2.018
  &0.0\%
  &NA
  &$10.00\times10^6$ 
  &$10.0\times 10^6$
      \\  \hline
       \textbf{CS} \cite{sheikholeslami2016hybrid}
   &10
  &2.036
  &0.0\%
  &2.079
  &$4.50\times10^6$ 
  &$5.0\times 10^6$
      \\  \hline
      \textbf{GAs} \cite{bi2015improved}
   &10
  &2.061
  &0.0\%
  &NA
  &NA 
  &$2.00\times 10^6$
   \\  \hline
      \textbf{GENOME} \cite{reca2006genetic}
   &10
  &2.302
  &0.0\%
  &2.334
  &$10.00\times 10^6$ 
  &$10.0\times 10^6$
      \\  \hline
      \textbf{CMA-ES(Continuous) }
  &10
  &
  &
  &
  &
  &
   \\
  
  $\lambda=200$& 
  &1.895
  &0.0\%
  &1.900
  &$0.84\times10^6$
  &$2\times 10^6$
   \\
  $\lambda=500$&
  &1.899
  &0.0\%
  &1.906
  &$0.98\times 10^6$
  &$2\times 10^6$
   \\
  \hline
   \textbf{CMA-ES(Discrete) }
  &10
  & 
  &
  &
  &
  &
   \\
  
  $\lambda=200$& 
  &1.974
  &0.0\%
  &1.990
  &$0.56\times10^6$
  &$2\times 10^6$
   \\
  $\lambda=500$&
  &1.961
  &0.0\%
  &1.971
  &$0.69 \times 10^6$
  &$2\times 10^6$
   \\
  \hline
   \textbf{CM$A_{ES}$-G$S_U$ }
  &10
  & 
  &
  &
  & 
  &
    \\ 
    $\lambda=200$&
  &1.936
  &0.0\%
  &1.943
  &$0.86\times10^6$
  &$2\times 10^6$
   \\
  $\lambda=500$&
  &1.937
  &0.0\%
  &1.942
  &$1.10 \times 10^6$
  &$2\times 10^6$
   \\
  \hline
   \textbf{CM$A_{ES}$-G$S_U$-G$S_D$ }
  &10
  & 
  &
  &
  &
  &
    \\ 
    $\lambda=200$&
  &1.9245
  &0.0\%
  &{1.9259}
  &$0.89\times10^6$
  &$2\times 10^6$
   \\
  $\lambda=500$&
  &1.9243
  &0.0\%
  &{1.9249}
  &$1.15\times10^6$
  &$2\times 10^6$
   \\
   \hline
\end{tabular}
}
\label{summary_per_BN}
\end{table*}
%---------------------------------------------
\subsection{Case Study 5: Balerma (BN) }
The fifth case study is the Balerma Network (BN), which is an irrigation WDS established in the province of Almeria (Spain) \cite{reca2006genetic}.  Its components are four reservoirs,  454 pipes, eight loops and 443 demand nodes. There are 10 PVC commercial pipes diameter sizes 125 to 600 mm. Therefore, the search space is $10^{454}$, which is considerably larger than the previous three case studies in this paper, and it is categorized as a large-scale optimization problem. The minimum required nodal pressure is 20 m. Pipe costs and other details are given in the reference \cite{reca2006genetic}.

The current best layout of the BN, which is found by Zheng et al. is at \euro 1.923 million. This functional design is achieved by a combination of the DE and nonlinear programming (NLP-DE). As demonstrated by the results from the Table \ref{summary_per_BN}, The average performance of the proposed hybrid framework is clearly better than all previous methods in terms of quality, efficiency and the convergence rate, mainly where CMA-ES is applied by the continuous decision variables. The best-introduced solution cost of the continuous CMA-ES is \euro 1.895 million ($\lambda=500$). The main objective of the study is evaluating the performance of the hybrid framework with discrete pipe sizes (commercialized), so the discrete results of CM$A_{ES}$-G$S_U$ and CM$A_{ES}$-G$S_U$-G$S_D$ are reported in the Table \ref{summary_per_BN} too.

According to the results of Table \ref{summary_per_BN}, except the excellent BN designs which are found by the CMA-ES (Continuous), we can see the average discrete best-founded BN designs overcome the existing methods. The saving rate of computational cost is 60\%. This feature illustrates the high ability of exploitation of both CM$A_{ES}$-G$S_U$ and  CM$A_{ES}$-G$S_U$-G$S_D$ methods and indicates that the proposed optimization framework is able to locate reasonable quality solutions with substantially developed computational effectiveness when faced with the large-scale WDS. It is noted that (Table \ref{summary_per_BN}), in terms of the success rate, the proposed hybrid framework can not overcome the NLP-DE2.

% Figure \ref{fig:BN_convergence_quality} represents the distribution of 10 obtained solutions by the best previous EAs (NLP-DE1, NLP-DE2 and DE3 \cite{c26}) and the new proposed hybrid framework (CMA-ES, CM$A_{ES}$-G$S_U$ and CM$A_{ES}$-G$S_U$-G$S_D$). 
It is crystal clear that the proposed hybrid method achievements are placed overall lower cost BN layouts compared to total previous methods with less computational budgets. Meanwhile, In terms of convergence speed, the discrete CMA-ES is located in the highest rank when the average evaluation number is just $0.56 \times 10^6$ and the best solution cost (near-optimum) is with a value of \euro 1.961 million. It is right that the quality of the proposed solutions of the discrete CMA-ES is not the best, while it is able to converge to the semi-optimal solutions 18, 8, 18, 16-fold faster than the HS \cite{geem2009particle}, CS \cite{sheikholeslami2016hybrid}, GENOME \cite{reca2006genetic} and DE3 \cite{zheng2011combined} respectively. 

The CM$A_{ES}$-G$S_U$ converged slightly slower than the discrete CMA-ES, but the quality of their outcomes are better and feasible (possible pipe sizes) as seen in Figure \ref{fig:BN_boxplot_quality}. Finally, the third part of the hybrid framework is evaluated for analyzing its impact on improving the results of the previous step (CM$A_{ES}$-G$S_U$). Where the dimension of the problem (BN) is high that leads to high value for branching factor of the tree structure of search space,  it is recommended that applying the Downward Greedy Search ($GS_D$) can be efficient because of its computational complexity and memory usage. One of the most important the $GS_D$ advantages is reducing the cost of BN layout by 0.6\% and 0.68\% ($\lambda$=200, 500) respectively on average with spending a few more percentages of the computational budgets.

For comparing the robustness of the hybrid framework convergence rate for the large-scale Balerma network with the best previous methods, Figure \ref{fig:BN_boxplot_quality} is drawn. Except for the performance of discrete CMA-ES, all three parts of the proposed method performed well. The best cost of the BN design that is found by the third step of the hybrid framework is \euro 1.9243 million that shows a worthwhile contribution and development versus the best previous methods as the second rank.      

\section{Conclusion}
The computational complexity is remarkably high in interpreting the optimization of the WDS problem. This optimization problem relates to an accumulation of inherently intractable problems referred to as NP-complete problems with nonlinear constraints.
A new hybrid optimization framework is proposed for optimizing the WDSs designs in this paper. The optimization process of the new framework is divided into three phases including 

1. Applying a robust and self-adaptive EAs called CMA-ES with the different pipe diameter sizes scenarios such as continuous, discrete(interval=1 inch) and possible (commercialized).

2. Carrying out an Upward Greedy Search ($GS_U$) for fixing up the violation of the nodal head constraints that are being optimized; and

3. Removing the extra imposed the cost of pipes which are related to the second step by utilizing a Downward Greedy Search ($GS_D$).

According to the optimization results, it can be shown that the proposed new combined framework has higher convergence characteristics for the large-scale network considerably. For both the NYTP and NYTP2 case studies, the hybrid approach is able to find new continuous feasible designs which are the cheapest ones at \$38.00 and \$72.00 million, and also the current best-known solutions are found more frequently (100\%) and more efficiently compared with other previous techniques. Besides, the best-introduced design of the HP case study is achieved by the proposed method too. Where the performance of CM$A_{ES}$-G$S_U$-G$S_D$ is better than the standard CMA-ES and some of the proposed methods based on the obtained solutions.

For the BN case study, the proposed new framework discovers and introduces the new cheap feasible design at \euro 1.9243 million. It shows a relatively substantial average design improvement in terms of both efficiency and robustness. The compatible superior achievement of the proposed method on four of five case studies demonstrates that the proposed hybridization is entirely satisfactory for the minimization of the WDSs cost.

The adaptability and extensibility are two important benefits of The Hybrid framework to optimize more complex WDS designs such as developing with pumps or other network components. Additionally, although the Hybrid framework utilizes EPANET as a fitness function, other hydraulic models can be incorporated. 
 \afterpage{%
     \clearpage% Flush earlier floats (otherwise order might not be correct)
    \thispagestyle{empty}% empty page style (?)
     \begin{landscape}% Landscape page
         \centering % Center table
         \begin{table}[b]
         \tiny
        % \scalebox{0.55}{
  %\centering
 \caption{The review of the best proposed solutions for NYTP (The best continuous solution is \$38,001,951 which is achieved by CMA-ES ), LP=Linear Programming, PE=Partial Enumeration, BC=Branched Configuration, DC= Decomposition, NLP=Non-Linear Programming, IGA= Improved Genetic Algorithm, GA-O= Optimizer GA, SFL= Shuffled Frog Leaping.{(CMAES solution is feasible in terms of nodal pressure with the continuous pipe diameter) }
  }
  \label{table_review_NYTP}
 \setlength{\tabcolsep}{4pt}

 \begin{tabular}{?p{0.8cm}|p{1cm}|p{0.9cm}|p{0.9cm}|p{0.9cm}| p{1.1cm}|p{0.9cm}|p{1.2cm}|p{1cm}|p{1.2
 cm}|p{1cm}|p{1.2cm}|p{1.2cm}|p{1.2cm}|p{1cm}|p{1cm}|p{1.2cm}|p{1cm}?
 }
 \hlineB{4}
 \textbf{Pipe/ Authors} &\textbf{Schaake, Lai} & \textbf{Quindry et al.} & \textbf{Gessler} & \textbf{Bhave} &\textbf{Morgan, Goulter} & \textbf{Kessler} & \textbf{Fujiwara
 Khang} & \textbf{Dandy et al.\cite{dandy1996improved}} &\textbf{Savic, Walters} & \textbf{Savic, Walters} & \textbf{Lippai et al.} & \textbf{Lippai et al.}&\textbf{Eusuff et al.} & \textbf{Maier et al.\cite{maier2003ant}} & \textbf{Sedki, Quazar} & \textbf{Aghdam et al.} & \textbf{Our solution}   \\ \hline
 \hline
% %\setlength{\tabcolsep}{10pt}
 \textbf{Year} &\textbf{1969} & \textbf{1981} & \textbf{1982} & \textbf{1985} &\textbf{1985} & \textbf{1988} & \textbf{1990} & \textbf{1996} &\textbf{1997} &\textbf{1997}& \textbf{1999} & \textbf{1999} & \textbf{2003}&\textbf{2003} & \textbf{2012} & \textbf{2014} & \textbf{2019}  \\ \hline
 \hlineB{4}
% %\setlength{\tabcolsep}{10pt}
 \textbf{Feasibility} &\textbf{Feasible} & \textbf{Feasible} & \textbf{Feasible} & \textbf{Feasible} &\textbf{
 Feasible} & \textbf{Clearly infeasible} & \textbf{Clearly
 infeasible} & \textbf{Feasible} &\textbf{Infeasible} & \textbf{Feasible} & \textbf{Very Slightly
 infeasible} & \textbf{Slightly infeasible}&\textbf{Very Slightly infeasible} & \textbf{Feasible} &\textbf{Feasible}& \textbf{Slightly infeasible} & \textbf{Feasible}  \\ \hlineB{4}
 \addlinespace[0.1cm]
   1& 52.02  &0.0  &0 &0.0 &0   &0.0  & 0.0&0 & 0  &0  &0 &0 & 0  &0  & 0& 0&0 \\ \hline
   \addlinespace[0.1cm]
   2&49.90   &0.0  & 0& 0.0& 0   & 0.0 &0.0 &0 & 0   &0  &0 &0 &0    &0  &0 &0&0\\ \hline
   \addlinespace[0.1cm]
   3&63.41   &0.0  & 0& 0.0&  0 &0.0  &0.0 &0 &0   &0  &0 &0 &0   & 0 &0 &0&0  \\ \hline
   \addlinespace[0.1cm]
   4&55.59   &0.0  & 0& 0.0&   0 & 0.0 &0.0 & 0&  0  &0  &0 &0 & 0   & 0 &0 &0&0\\ \hline
   \addlinespace[0.1cm]
   5&57.25   &0.0  & 0& 0.0&   0& 0.0 &0.0 & 0& 0  & 0 &0 &0 &0   & 0 &0 &0&0  \\ \hline
   \addlinespace[0.1cm]
   6&59.19   &0.0  & 0& 0.0&    0& 0.0 &73.62 &0 & 0   &0  &0 &0 & 0   & 0 & 0&0&0\\ \hline
   \addlinespace[0.1cm]
   7&59.06   &0.0  & 100& 0.0&   144& 0.0 &0.0 & 0& 108  &0  &132 &124 & 132  & 144 &144 &0&118.99  \\ \hline
   \addlinespace[0.1cm]
   8&54.95   &0.0  & 100& 0.0&    0& 0.0 & 0.0& 0& 0   & 0 &0 &0 & 0   & 0 & 0&0&0\\ \hline
   \addlinespace[0.1cm]
   9&0.0   &0.0  & 0& 0.0& 0  & 0.0 &0.0 &0 &  0 & 0 &0 &0 &0   &0  & 0& 0& 0 \\ \hline
   \addlinespace[0.1cm]
   10&0.0   &0.0  & 0& 0.0&  0  & 0.0 & 0.0& 0&  0  & 0 &0 &0 & 0   & 0 &0 &0 &0\\ \hline
   \addlinespace[0.1cm]
   11&116.21   &119.02  &0 &0.0 &  0 & 0.0 &0.0 &0 & 0  & 0 &0 &0 & 0  &0  &0 & 0&0 \\ \hline
   \addlinespace[0.1cm]
   12&125.25   & 134.39 & 0&0.0 &  0  &0.0  & 0.0&0 &  0  & 0 & 0& 0& 0  & 0 &0 &0 &0\\ \hline
   \addlinespace[0.1cm]
   13&126.87   &132.49  & 0&0.0 & 0  & 0.0 & 0.0& 0& 0  & 0 &0 &0 &0   &0  & 0& 0&0 \\ \hline
   \addlinespace[0.1cm]
   14&133.07   &132.87  & 0&0.0 & 0   & 0.0 & 0.0&0 &  0  &  0& 0 &0 & 0   & 0 & 0&0 &0\\ \hline
   \addlinespace[0.1cm]
   15&126.52   &131.37  & 0& 136.43& 0  & 156.11 & 0.0& 120&  0 &144  & 0& 0& 0  &0  & 0&  96 &0\\ \hline
   \addlinespace[0.1cm]
   16&19.52   &19.26  &100 &87.37&  96  & 72.00 & 99.01& 84& 96   & 84 & 96& 96& 96   & 96 & 96&96 &99.97\\ \hline
   \addlinespace[0.1cm]
   17&91.83   &91.71  &100 &99.23&  96 & 96.60 &98.75 &96 & 96  &96  &96 &96 & 96  & 96 & 96& 96& 99.28 \\ \hline
   \addlinespace[0.1cm]
   18&72.76   &72.76  &80 & 78.17&  84  &78.00  &78.97 & 84&  84  &84  &84 & 84& 84   &84  &84 &84 &79.09\\ \hline
   \addlinespace[0.1cm]
   19&72.61   &72.64  &60 & 54.40& 60  &59.78  &83.82 &72 & 72  & 72 &72 & 72& 72  & 72 &72 & 72& 75.06 \\ \hline
   \addlinespace[0.1cm]
   20&0.0   &0.0  &0 & 0.0& 0   & 0.0 &0.0 &0 &  0  &0  &0 &0 & 0   & 0 & 0&0 &0\\ \hline
   \addlinespace[0.1cm]
   21&54.82   &54.97  &80 &81.50 & 84  &72.27  &66.59 & 72& 72  & 72 &72 &72 & 72  & 72 &72 & 72 & 70.60\\ \hlineB{4} \hline
   \addlinespace[0.1cm]
   Node 16 Excess
   &+1.099  
   &+1.0266 
   &+0.3902
   &+0.9154
   &+1.6220
   &-1.4211
   &-0.8765
   &+0.5894 
   &-0.2061
   &+1.1849 
   &-0.0021 
   &-0.0634
   &-0.0021
   &+0.0771 
   &+0.0771 
   &+0.0194
   &\textbf{0.0000}
      \\ \hline
    Node 17 Excess   
   &+1.0704  
   &+0.9061 
   &+0.3552
   &+0.6290
   &+0.0447
   &+0.2879
   &-0.7332
   &+0.1099
   &-0.2174
   &+0.7020 
   &-0.0116
   &-0.0734
   &-0.0116 
   &+0.0684 
   &+0.0684 
   &-0.0036
   & \textbf{0.0000}
     \\ \hline
   Node 19 Excess
   &+1.2217  
   &+1.1083 
   &+0.9317 
   &+1.0296 
   &+0.0667
   &+0.2911
   &-0.8979
   &+0.7782
   &-0.1977
   &+1.4133 
   &-0.0164
   &-0.0709
   &-0.0164  
   &+0.0540 
   &+0.0540 
   &+0.1714 
   & \textbf{0.0000} 
   \\ \hlineB{4} \hline
   Total Cost(\$M)&78.09   &63.58  &41.8 &40.18 & 39.20   &38.96  &38.30 & 38.80& 37.13   & 40.42 & 38.13&  37.83   &38.13  & 38.64& 38.64&38.52 & {\textbf{38.00}}\\ \hline
   \addlinespace[0.1cm] \hlineB{4}
     Methods&LP   &LP  &PE & BC& LP   & DC &NLP &I-GA &GA    & GA & GA-O &GA-O & SFL   & ACO &PSO &AMPSO&{\textbf{CMA-ES}}\\ \hline
 
 \end{tabular}

 % \label{NYTP_review}
 \end{table}
     \end{landscape}
     \clearpage% Flush page
 }

%---------------------
% \section{CONCLUSIONS}
 \afterpage{%
     \clearpage% Flush earlier floats (otherwise order might not be correct)
    \thispagestyle{empty}% empty page style (?)
     \begin{landscape}% Landscape page
         \centering % Center table
         \begin{table*}[b]
         \tiny
      %  \scalebox{0.55}{
       
  %\centering
 \caption{The summery of the best proposed solutions for {Hanoi Network} (The best discovered solution is {\$6.081(M\$)}) }
  
  \label{table_review_HP}
 \setlength{\tabcolsep}{2pt}

 \begin{tabular}{?p{1.3cm}?p{1.1cm}|p{0.7cm}?p{1.1cm}|p{0.8cm}? p{1.2cm}|p{1.0cm}?p{1.2cm}|p{1cm}?p{1.1cm}|p{1cm}?p{1.2cm}|p{1.2cm}?p{1.2cm}|p{1cm}?p{1.1cm}|p{1.0cm}?}
 \hlineB{4}
 \textbf{ Pipe/ Authors} 
 &\multicolumn{2}{|c|}{{Fujiwara}}
 &\multicolumn{2}{|c|}{{Savic \& Walters}} 
 &\multicolumn{2}{|c|}{{Cunha \& Sousa}}   
 & \multicolumn{2}{|c|}{{Wu et al.}} 
 &\multicolumn{2}{|c|}{{Zecchin \& Simpson}} 
 &\multicolumn{2}{|c|}{{Zecchin \& Simpson}} 
 &\multicolumn{2}{|c|}{{Sedki\& Ouazar}} 
 &\multicolumn{2}{|c|}{{Our solution}}
  \\ \hline
 \hline
% %\setlength{\tabcolsep}{10pt}
 \textbf{Year} 
 &\multicolumn{2}{|c|}{\textbf{1990}}  
 &\multicolumn{2}{|c|}{\textbf{1997}}
 &\multicolumn{2}{|c|}{\textbf{1999}}
 &\multicolumn{2}{|c|}{\textbf{2001}}
 &\multicolumn{2}{|c|}{\textbf{2005}}
 &\multicolumn{2}{|c|}{\textbf{2006}}
 &\multicolumn{2}{|c|}{\textbf{2012}}
 &\multicolumn{2}{|c|}{\textbf{2019}}
 \\ \hline
 \hlineB{4}
% %\setlength{\tabcolsep}{10pt}
 \textbf{Feasibility} 
 &\multicolumn{2}{|c|}{\textbf{Clearly infeasible}}  
 &\multicolumn{2}{|c|}{\textbf{Infeasible}}  
 & \multicolumn{2}{|c|}{\textbf{Infeasible}}   
 & \multicolumn{2}{|c|}{\textbf{Feasible} }
 & \multicolumn{2}{|c|}{\textbf{Feasible} }
 & \multicolumn{2}{|c|}{\textbf{Feasible}}
 & \multicolumn{2}{|c|}{\textbf{Feasible} }
 & \multicolumn{2}{|c|}{\textbf{Feasible}}
  \\ \hlineB{4}
 \addlinespace[0.1cm]
 \textbf{\#} &Diameter&Head& Diameter&Head  & Diameter&Head &  Diameter&Head  & Diameter&Head  &  Diameter&Head & Diameter&Head  & Diameter&Head  \\ \hlineB{3}
 \addlinespace[0.1cm]
   1
   &40 &100 &40 &100 &40   &100  & 40&100 & 40  &100  &40 &100 & 40  &100  & 40&100 \\ \hline
   
   2
   &40 &97.14 &40 &97.14 &40   &97.14  & 40&97.14 & 40  &97.14  &40 &97.14 & 40  &97.14  & 40&97.14   \\ \hline
   
   3
   &38.8 &61.67 &40 &61.67 &40   &61.67 & 40&61.67 & 40  &61.67  &40 &61.67 & 40  &61.67  &40 &61.67  \\ \hline
   
   4
   &38.7 &56.17 &40 &56.88 &40   &56.87  & 40&57.22 & 40  &57.63  &40 &57.08& 40  &56.92  & 40&57.17  \\ \hline
   
   5
   &37.8 &49.27 &40 &50.94 &40   &50.92  & 40&51.70 & 40  &52.63  &40 &51.38 & 40  &51.03  & 40&51.61 \\ \hline
   
   6
   &36.3 &41.1 &40 &44.68 &40   &44.64  & 40&45.93 & 40  &47.45  &40 &45.40 & 40  &44.81  & 40&45.77   \\ \hline
  
   7
   &33.8   &38.77  & 40& 43.21&   40& 43.16 &40 & 44.59& 40  &46.27  &40 &44.00 & 40  & 43.35 &37.68 &44.42   \\ \hline
  
   8
   &32.8   &34.84  & 40& 41.45&    40& 41.39 & 40& 43.03& 40   &44.96 &40 &42.36 & 40   &41.62 & 36.29&42.30 \\ \hline
  
   9
   &31.5   &31.2  & 40& 40.04& 40  &39.98&40 &41.81 &  30 & 43.95 &40 &41.06 &40   &40.23  & 34.84&40.30  \\ \hline
   
   10
   &25   &\textit{27.94}  & 30& 38.2&  30  & 38.93 & 30& 40.92&  30  & 41.07&30 &40.11 &30   & 39.20 &29.17 &38.52  \\ \hline
  
   11
   &23   &\textit{24.15} &24 &37.44 &  24 &37.37 &24 &39.36 & 24  & 39.51 &24 &38.55 &24  &37.64  &26.83 &36.73 \\ \hline
  
   12
   &20.2   & \textit{19.93} & 24&34.01 &  24  &33.94 & 24&35.94 &  24  & 36.08 & 24& 35.12&24  & 34.22 &23.42 &34.74   \\ \hline
  
   13
   &19   &\textit{10.19 } & 20&\textit{29.80} & 20  & \textit{29.74} & 16& 31.73& 12  & 31.87 &20 &30.91&20   &30.01  & 16.67&30.00  \\ \hline
  
   14
   &14.5   &\textit{23.29}  & 16&35.13 & 16   & 35.00 & 12&34.22 &  12  &  31.31& 12 &37.21 &16   & 35.52 & 12.35&32.58  \\ \hline
  
   15
   &12   &\textit{20.45}  & 12& 33.14& 12  & 32.95 & 12& 31.99&  12 &31.51  &12& 32.89& 12  &33.72  & 12.00&30.15   \\ \hline
  
   16
   &19.9   &\textit{18.2}  &12 &30.23&  12 & \textit{29.87} & 12& 31.96& 24   & 34.86 & 12& 32.16& 12   & 31.30 & 12.00&30.00  \\ \hline
  
   17
   &23.1   &37.01  &16 &30.33& 16 & 30.03 &20 &44.73 & 20 &38.38  &20 &41.36 & 16  & 33.41 & 17.70&37.24 \\ \hline
  
   18
   &26.6   &50.72  &20 & 43.97&  20  &43.87  &24 & 52.54&  40  &56.13  &24 & 48.55&24  &49.93  &22.32 &49.60  \\ \hline
  
   19
   &26.8   &58.04  &20 & 55.58& 20  &55.54  &24 &58.54 & 24  & 56.9 &20 & 54.33& 20  & 55.09 &22.49 &57.62  \\ \hline
   
   20
   &35.2   &47.15  &40 & 50.44& 40   & 50.49 &40 &50.4 &  40  &51.1 &40 &50.61 &40   & 50.61 & 40&50.28  \\ \hline
  
   21
   &16.4   &\textit{22.57}  &20 &41.09 & 20  &41.14  &20 & 41.05&20  & 41.75 &20 &41.26 & 20  & 41.26 &17.74 &33.50 \\ \hline
22
& 12  &\textit{17.41} &12 &35.93 &12   &35.98 & 12&35.88 &12  &36.58  &12 &36.1 & 12  &36.1  & 12.99&30.00   \\ \hline
23
& 29.5  &31.91  &40 &44.21 &40   &44.3  & 40&44.13 & 40  &45.41  &40 &44.53 & 40  &44.53  & 37.97&42.08  \\ \hline
24
& 19.3  &\textit{20.02}  &30 &38.9 &30   &38.57  & 30&38.66 & 30  &41.33  &30 &39.39 & 30  &38.93  & 32.66&38.21  \\ \hline
25
& 16.4  &\textit{12.39 } &30 &35.56&30   &34.86  & 30&35.17 & 24  &34.36  &30 &36.19 & 30  &35.34  & 30.26&34.55 \\ \hline
26
& 12  &\textit{13.93}  &20 &31.53 &20   &30.95  & 24&33.48 & 20  &31.73  &20 &32.55 & 20  &31.7  & 20.64&30.79   \\ \hline
27
& 20  &\textit{15.08}  &12 &30.10 &12   &\textit{29.66} & 12&31.91 &12  &31.65  &12 &31.61 & 12  &30.76  & 14.49&30.00  \\ \hline
28
& 22  &\textit{24.36 } &12 &35.5 &12   &38.66 & 12&36.22 &12   &40.93 &12 &35.90 &12  &38.94  & 12.00&36.10  \\ \hline
29
& 18.9  &\textit{15.12}  &16&30.75 &16   &\textit{29.72} & 16&32.17 & 20  &31.89  &16 &31.23 &16  &30.13  & 16.17&30.00  \\ \hline
30
& 17.1  &\textit{9.88} &16 &\textit{29.73}&12   &\textit{29.98}  &16&31.72 & 16  &30.20  &16 &30.29 & 12  &30.42 & 13.54&30.03  \\ \hline
31
& 14.6  &\textit{9.83 } &12 &30.19 &12   &30.26  & 12&31.87 & 16  &30.24  &12 &30.77 & 12  &30.70  & 12.00&30.28 \\ \hline
32
& 12  &\textit{9.88  }&12 &31.44 &16   &32.72  & 16&33.37 & 12  &31.42  &12 &32.4 & 16  &33.18  & 15.84&31.86   \\ \hline
33
& 12  &  &16 & &16   &  & 16& & 12  &  &16 & & 16  &  & 17.02& \\ \hline
34
& 19.5  &  &20 & &24   &  & 24& & 20  &  &20 & &24  & & 22.67&  \\ \hline\hline
   \addlinespace[0.1cm]
   \textbf{Total Cost}(\$M)
   & \multicolumn{2}{|c|}{\textbf{5.351}}   
   & \multicolumn{2}{|c|}{\textbf{6.072}} 
   & \multicolumn{2}{|c|}{\textbf{6.056}} 
   & \multicolumn{2}{|c|}{\textbf{6.183}} 
   & \multicolumn{2}{|c|}{\textbf{6.367}} 
   & \multicolumn{2}{|c|}{\textbf{6.134}} 
   & \multicolumn{2}{|c|}{\textbf{6.081}} 
   & \multicolumn{2}{|c|}{{\textbf{5.959}}} 
   \\ \hline\hline
   \addlinespace[0.01cm]
     \textbf{Methods}
     &  \multicolumn{2}{|c|}{\textbf{NLPG-ILOS \cite{fujiwara1990two}}} 
     & \multicolumn{2}{|c|}{\textbf{GANET \cite{savic1997genetic}}}
     & \multicolumn{2}{|c|}{\textbf{SA \cite{cunha1999water}}}
     &\multicolumn{2}{|c|}{\textbf{GA \cite{wu2001using}}}
     & \multicolumn{2}{|c|}{\textbf{ACO \cite{zecchin2005parametric}}}
     &\multicolumn{2}{|c|}{\textbf{Two-ACO \cite{zecchin2006application}}}
     &\multicolumn{2}{|c|}{\textbf{PSO-DE \cite{sedki2012hybrid}}}
     &\multicolumn{2}{|c|}{{\textbf{CMA-ES,$\lambda=200$}}}\\ \hline
 
 \end{tabular}
 %}

 % \label{NYTP_review}
 \end{table*}
     \end{landscape}
     \clearpage% Flush page
 }

%-------------------------
 \section*{Acknowledgements}
 We would like to offer our special thanks to Dr. Markus Wagner, Dr. Holger Maier and Dr. Mengning Qiu for their valuable and constructive suggestions.

\bibliographystyle{unsrt}  

\bibliography{references} 

\begin{thebibliography}{10}

\bibitem{alperovits1977design}
Elyahu Alperovits and Uri Shamir.
\newblock Design of optimal water distribution systems.
\newblock {\em Water resources research}, 13(6):885--900, 1977.

\bibitem{fujiwara1987modified}
O~Fujiwara, B~Jenchaimahakoon, and NCP Edirishinghe.
\newblock A modified linear programming gradient method for optimal design of
  looped water distribution networks.
\newblock {\em Water Resources Research}, 23(6):977--982, 1987.

\bibitem{fujiwara1990two}
Okitsugu Fujiwara and Do~Ba Khang.
\newblock A two-phase decomposition method for optimal design of looped water
  distribution networks.
\newblock {\em Water resources research}, 26(4):539--549, 1990.

\bibitem{savic1997genetic}
Dragan~A Savic and Godfrey~A Walters.
\newblock Genetic algorithms for least-cost design of water distribution
  networks.
\newblock {\em Journal of water resources planning and management},
  123(2):67--77, 1997.

\bibitem{wu2001using}
Zheng~Y Wu, Paul~F Boulos, Chun~Hou Orr, and Jun~Je Ro.
\newblock Using genetic algorithms to rehabilitate distribution systems.
\newblock {\em Journal-American Water Works Association}, 93(11):74--85, 2001.

\bibitem{bi2015improved}
Weiwei Bi, Graeme~C Dandy, and Holger~R Maier.
\newblock Improved genetic algorithm optimization of water distribution system
  design by incorporating domain knowledge.
\newblock {\em Environmental Modelling \& Software}, 69:370--381, 2015.

\bibitem{maier2003ant}
Holger~R Maier, Angus~R Simpson, Aaron~C Zecchin, Wai~Kuan Foong, Kuang~Yeow
  Phang, Hsin~Yeow Seah, and Chan~Lim Tan.
\newblock Ant colony optimization for design of water distribution systems.
\newblock {\em Journal of water resources planning and management},
  129(3):200--209, 2003.

\bibitem{geem2009particle}
Zong~Woo Geem.
\newblock Particle-swarm harmony search for water network design.
\newblock {\em Engineering Optimization}, 41(4):297--311, 2009.

\bibitem{zecchin2006application}
Aaron~C Zecchin, Angus~R Simpson, Holger~R Maier, Michael Leonard, Andrew~J
  Roberts, and Matthew~J Berrisford.
\newblock Application of two ant colony optimisation algorithms to water
  distribution system optimisation.
\newblock {\em Mathematical and computer modelling}, 44(5-6):451--468, 2006.

\bibitem{zecchin2007ant}
Aaron~C Zecchin, Holger~R Maier, Angus~R Simpson, Michael Leonard, and John~B
  Nixon.
\newblock Ant colony optimization applied to water distribution system design:
  Comparative study of five algorithms.
\newblock {\em Journal of Water Resources Planning and Management},
  133(1):87--92, 2007.

\bibitem{zheng2017adaptive}
Feifei Zheng, Aaron~C Zecchin, Jeffery~P Newman, Holger~R Maier, and Graeme~C
  Dandy.
\newblock An adaptive convergence-trajectory controlled ant colony optimization
  algorithm with application to water distribution system design problems.
\newblock {\em IEEE Transactions on Evolutionary Computation}, 21(5):773--791,
  2017.

\bibitem{cunha1999water}
Maria da~Conceicao Cunha and Joaquim Sousa.
\newblock Water distribution network design optimization: simulated annealing
  approach.
\newblock {\em Journal of water resources planning and management},
  125(4):215--221, 1999.

\bibitem{lin2007scatter}
Min-Der Lin, Yu-Hsin Liu, Gee-Fon Liu, and Chien-Wei Chu.
\newblock Scatter search heuristic for least-cost design of water distribution
  networks.
\newblock {\em Engineering Optimization}, 39(7):857--876, 2007.

\bibitem{eusuff2003optimization}
Muzaffar~M Eusuff and Kevin~E Lansey.
\newblock Optimization of water distribution network design using the shuffled
  frog leaping algorithm.
\newblock {\em Journal of Water Resources planning and management},
  129(3):210--225, 2003.

\bibitem{montalvo2008particle}
Idel Montalvo, Joaqu{\'\i}n Izquierdo, Rafael P{\'e}rez, and Michael~M Tung.
\newblock Particle swarm optimization applied to the design of water supply
  systems.
\newblock {\em Computers \& Mathematics with Applications}, 56(3):769--776,
  2008.

\bibitem{aghdam2014design}
Kazem~Mohammadi Aghdam, Iraj Mirzaee, Nader Pourmahmood, and
  Mohammad~Pourmahmood Aghababa.
\newblock Design of water distribution networks using accelerated momentum
  particle swarm optimisation technique.
\newblock {\em Journal of Experimental \& Theoretical Artificial Intelligence},
  26(4):459--475, 2014.

\bibitem{suribabu2010differential}
CR~Suribabu.
\newblock Differential evolution algorithm for optimal design of water
  distribution networks.
\newblock {\em Journal of Hydroinformatics}, 12(1):66--82, 2010.

\bibitem{zheng2012performance}
Feifei Zheng, Angus~R Simpson, and Aaron Zecchin.
\newblock A performance comparison of differential evolution and genetic
  algorithm variants applied to water distribution system optimization.
\newblock In {\em World Environmental and Water Resources Congress 2012:
  Crossing Boundaries}, pages 2954--2963, 2012.

\bibitem{zheng2012self}
Feifei Zheng, Aaron~C Zecchin, and Angus~R Simpson.
\newblock Self-adaptive differential evolution algorithm applied to water
  distribution system optimization.
\newblock {\em Journal of Computing in Civil Engineering}, 27(2):148--158,
  2012.

\bibitem{zheng2013coupled}
Feifei Zheng, Angus~R Simpson, and Aaron~C Zecchin.
\newblock Coupled binary linear programming--differential evolution algorithm
  approach for water distribution system optimization.
\newblock {\em Journal of Water Resources Planning and Management},
  140(5):585--597, 2013.

\bibitem{zheng2011combined}
Feifei Zheng, Angus~R Simpson, and Aaron~C Zecchin.
\newblock A combined nlp-differential evolution algorithm approach for the
  optimization of looped water distribution systems.
\newblock {\em Water Resources Research}, 47(8), 2011.

\bibitem{sedki2012hybrid}
A~Sedki and Driss Ouazar.
\newblock Hybrid particle swarm optimization and differential evolution for
  optimal design of water distribution systems.
\newblock {\em Advanced Engineering Informatics}, 26(3):582--591, 2012.

\bibitem{hansen2004evaluating}
Nikolaus Hansen and Stefan Kern.
\newblock Evaluating the cma evolution strategy on multimodal test functions.
\newblock In {\em International Conference on Parallel Problem Solving from
  Nature}, pages 282--291. Springer, 2004.

\bibitem{Skvorc:2019:GBO:3319619.3321996}
Urban \v{S}kvorc, Tome Eftimov, and Peter Koro\v{s}ec.
\newblock Gecco black-box optimization competitions: Progress from 2009 to
  2018.
\newblock In {\em Proceedings of the Genetic and Evolutionary Computation
  Conference Companion}, GECCO '19, pages 275--276, New York, NY, USA, 2019.
  ACM.

\bibitem{beyer2002evolution}
Hans-Georg Beyer and Hans-Paul Schwefel.
\newblock Evolution strategies--a comprehensive introduction.
\newblock {\em Natural computing}, 1(1):3--52, 2002.

\bibitem{hansen2014principled}
Nikolaus Hansen and Anne Auger.
\newblock Principled design of continuous stochastic search: From theory to
  practice.
\newblock In {\em Theory and principled methods for the design of
  metaheuristics}, pages 145--180. Springer, 2014.

\bibitem{hansen2006cma}
Nikolaus Hansen.
\newblock The cma evolution strategy: a comparing review.
\newblock In {\em Towards a new evolutionary computation}, pages 75--102.
  Springer, 2006.

\bibitem{neumann2007randomized}
Frank Neumann and Ingo Wegener.
\newblock Randomized local search, evolutionary algorithms, and the minimum
  spanning tree problem.
\newblock {\em Theoretical Computer Science}, 378(1):32--40, 2007.

\bibitem{neshat2018detailed}
Mehdi Neshat, Bradley Alexander, Markus Wagner, and Yuanzhong Xia.
\newblock A detailed comparison of meta-heuristic methods for optimising wave
  energy converter placements.
\newblock In {\em Proceedings of the Genetic and Evolutionary Computation
  Conference}, pages 1318--1325. ACM, 2018.

\bibitem{dandy1996improved}
Graeme~C Dandy, Angus~R Simpson, and Laurence~J Murphy.
\newblock An improved genetic algorithm for pipe network optimization.
\newblock {\em Water resources research}, 32(2):449--458, 1996.

\bibitem{zecchin2005parametric}
Aaron~C Zecchin, Angus~R Simpson, Holger~R Maier, and John~B Nixon.
\newblock Parametric study for an ant algorithm applied to water distribution
  system optimization.
\newblock {\em IEEE transactions on evolutionary computation}, 9(2):175--191,
  2005.

\bibitem{zheng2011optimal}
Feifei Zheng, Angus~R Simpson, and Aaron~C Zecchin.
\newblock Optimal rehabilitation for large water distribution systems using
  genetic algorithms.
\newblock {\em Proc., Australia Water Association (AWA) 2011 OzWater}, 2011.

\bibitem{reca2006genetic}
Juan Reca and Juan Mart{\'\i}nez.
\newblock Genetic algorithms for the design of looped irrigation water
  distribution networks.
\newblock {\em Water resources research}, 42(5), 2006.

\bibitem{eiger1994optimal}
Gideon Eiger, Uri Shamir, and Aharon Ben-Tal.
\newblock Optimal design of water distribution networks.
\newblock {\em Water resources research}, 30(9):2637--2646, 1994.

\bibitem{bolognesi2010genetic}
Andrea Bolognesi, Cristiana Bragalli, Angela Marchi, and Sandro Artina.
\newblock Genetic heritage evolution by stochastic transmission in the optimal
  design of water distribution networks.
\newblock {\em Advances in Engineering Software}, 41(5):792--801, 2010.

\bibitem{tolson2009hybrid}
Bryan~A Tolson, Masoud Asadzadeh, Holger~R Maier, and Aaron Zecchin.
\newblock Hybrid discrete dynamically dimensioned search (hd-dds) algorithm for
  water distribution system design optimization.
\newblock {\em Water Resources Research}, 45(12), 2009.

\bibitem{sheikholeslami2016hybrid}
Razi Sheikholeslami, Aaron~C Zecchin, Feifei Zheng, and Siamak Talatahari.
\newblock A hybrid cuckoo--harmony search algorithm for optimal design of water
  distribution systems.
\newblock {\em Journal of Hydroinformatics}, 18(3):544--563, 2016.

\end{thebibliography}

\end{document}